\documentclass[final,3p,times,twocolumn,authoryear]{elsarticle}
\usepackage{amssymb}

\usepackage{times}
\usepackage{epsfig}
\usepackage{graphicx}
\usepackage{amsmath}
\usepackage{amssymb}
\usepackage{mathtools}
\usepackage{standalone}
\usepackage{csquotes}
\usepackage[usenames, dvipsnames]{color}
\usepackage{color,soul}
\usepackage{tabularx}
\usepackage{amsmath}
\usepackage{nccmath}
\usepackage{color}
\usepackage[table,xcdraw]{xcolor}
\usepackage{float}

\newcommand{\ie}{\textit{i}.\textit{e}.}
\newcommand{\eg}{\textit{e}.\textit{g}.}

\journal{ISPRS Journal of Photogrammetry and Remote Sensing}

\begin{document}

\begin{frontmatter}

%% Title, authors and addresses

%% use the tnoteref command within \title for footnotes;
%% use the tnotetext command for theassociated footnote;
%% use the fnref command within \author or \address for footnotes;
%% use the fntext command for theassociated footnote;
%% use the corref command within \author for corresponding author footnotes;
%% use the cortext command for theassociated footnote;
%% use the ead command for the email address,
%% and the form \ead[url] for the home page:
%% \title{Title\tnoteref{label1}}
%% \tnotetext[label1]{}
%% \author{Name\corref{cor1}\fnref{label2}}
%% \ead{email address}
%% \ead[url]{home page}
%% \fntext[label2]{}
%% \cortext[cor1]{}
%% \address{Address\fnref{label3}}
%% \fntext[label3]{}

\title{A Fully Convolutional Network for Semantic Labeling of 3D Point Clouds}

%% use optional labels to link authors explicitly to addresses:
%% \author[label1,label2]{}
%% \address[label1]{}
%% \address[label2]{}

\author[add1]{Mohammed Yousefhussien}
\author[add2]{David J. Kelbe}
\author[add1]{Emmett J. Ientilucci}
\author[add1]{Carl Salvaggio}
\address[add1]{Rochester Institute of Technology, Chester F. Carlson Center for Imaging Science, Rochester, NY, USA}
\address[add2]{Oak Ridge National Laboratory, Geographic Information Science and Technology Group, Oak Ridge, TN, USA}
%\renewcommand\Authands{ and }
%\address{Rochester Institute of Technology}

\begin{abstract}
%% Text of abstract
When classifying point clouds, a large amount of time is devoted  to the process of engineering a reliable set of features which are then passed to a classifier of choice. Generally, such features -- usually derived from the 3D-covariance matrix -- are computed using the surrounding neighborhood of points. 
While these features capture local information, the process is usually time-consuming, and requires the application at multiple scales combined with contextual methods in order to adequately describe the diversity of objects within a scene. 
In this paper we present a 1D-fully convolutional network that consumes terrain-normalized points directly with the corresponding spectral data, if available, to generate point-wise labeling while implicitly learning contextual features in an end-to-end fashion.
Our method uses only the 3D-coordinates and three corresponding spectral features for each point. 
Spectral features may either be extracted from 2D-georeferenced images, as shown here for Light Detection and Ranging (LiDAR) point clouds, or extracted directly for passive-derived point clouds, \ie~ from muliple-view imagery. 
We train our network by splitting the data into square regions, and use a pooling layer that respects the permutation-invariance of the input points.
Evaluated using the ISPRS 3D Semantic Labeling Contest, our method scored second place with an overall accuracy of 81.6\%. 
We ranked third place with a mean F1-score of 63.32\%, surpassing the F1-score of the method with highest accuracy by 1.69\%. 
In addition to labeling 3D-point clouds, we also show that our method can be easily extended to 2D-semantic segmentation tasks, with promising initial results.

\end{abstract}

\begin{keyword}
%% keywords here, in the form: keyword \sep keyword
LiDAR \sep 3D-Labeling Contest \sep Deep Learning.
%% PACS codes here, in the form: \PACS code \sep code

%% MSC codes here, in the form: \MSC code \sep code
%% or \MSC[2008] code \sep code (2000 is the default)

\end{keyword}

\end{frontmatter}

%% \linenumbers

%% main text
\section{Introduction}
\label{sec:introduction}
The generation of dense, 3D-point clouds from overhead sensing systems is growing in scope and scale through processes such as Light Detection and Ranging (LiDAR), dense stereo- or multiview-photogrammetry and, in the computer vision domain, structure from motion (SfM). 
Despite the prevalence of 3D-point cloud data, however, automated interpretation and knowledge discovery from 3D-data remains challenging due to the irregular structure of raw point clouds. 
As such, exploitation has typically been limited to simple visualization and basic mensuration \citep{hackel2016fast}; or, the point cloud is rasterized onto a more tractable 2.5D- digital surface model (DSM) from which conventional image processing techniques are applied, \eg~ \citep{hug1997detecting, haala19983d}. 

In order to generate exploitation-ready data products directly from the point cloud, semantic classification is desired.
Similar to per-pixel image labeling, 3D-semantic labeling seeks to attribute a semantic classification label to each 3D-point. 
Classification labels, \eg~ vegetation, building, road, etc., can subsequently be used to inform derivative processing efforts, such as surface fitting, 3D modeling, object detection, and bare-earth extraction. 

\section{Related Work}
\label{sec:related_work}
Point cloud labeling algorithms can generally be grouped into two main categories. 
Section \ref{sec:direct_methods} describes ``Direct Methods'', which operate immediately on the point clouds themselves, and do not change the 3D-nature of the data. 
Section \ref{sec:indirect_methods} describes ``Indirect Methods'', which transform the input point cloud, \eg~ into an image or a volume, from which known semantic segmentation methods can then be applied. 
Finally, considering the relative trade-offs of these techniques, Section \ref{sec:contribution} proposes a novel approach with 7 specific contributions for semantic classification of point clouds. 

\subsection{Direct Methods}
\label{sec:direct_methods}
Direct methods assign semantic labels to each element in the point cloud based, fundamentally, on a simple point-wise discriminative model operating on point features. Such features, known as ``eigen-features'', are derived from the covariance matrix of a local neighborhood and provide information on the local geometry of the sampled surface, \eg~ planarity, sphericity, linearity \citep{lin2014eigen}. 
To improve classification, contextual information can explicitly be incorporated into the model. 
For example, \citet{Blomley16} used covariance features at multiple scales found using the eigenentropy-based scale selection method \citep{weinnman2014} and evaluated four different classifiers using the ISPRS 3D Semantic Labeling Contest\footnote{\url{https://goo.gl/fSK6Fy}}.
Their best-performing model used a Linear Discriminant Analysis (LDA) classifier in conjunction with various local geometric features.
However, scalability of this model was limited, due to the dependence upon various handcrafted features, and the need to experiment with various models that don't incorporate contextual features and require effort to tune.  

Motivated by the frequent availability of coincident 3D data and optical imagery, \citet{ramiya2014semantic} proposed the use of point coordinates and spectral data directly, forming a per-point vector of (X,Y,Z,R,G,B) components. 
Labeling was achieved by filtering the scene into ground and non-ground points according to \citet{axelsson2000generation}, then applying a 3D-region-growing segmentation to both sets to generate object proposals. 
Like \citet{Blomley16}, several geometric features were also derived, although specific details were not published.
Without incorporating contextual features, each segment/proposal was then classified into a selected set of five classes from the main ISPRS 3D Semantic Labeling Contest. 
 
% Moved the commented text below to later in the introduction
%which has the potential to improve results, but at the cost of computational efficiency. 
%On the other hand, by only using the point coordinates combined with simple three spectral values, we aim to show that without additional handcrafted features or 3D-segmentation, we were able to achieve near state-of-the-art results when tested on all nine classes presented in the main challenge. 
Several other methods were also reported in the literature such as the work by \citep{mallet2010} which classified full-waveform LiDAR data using a point-wise multiclass support vector machine (SVM), and \citep{Chehata09} who used random forests (RF) for feature detection and classification of urban scenes collected by airborne LiDAR.
The reader is referred to \citet{grilli2017review} for a review.
While simple discriminative models are well-established, they are unable to consider interactions between 3D points.

To allow for spatial dependencies between object classes by considering labels of the local neighborhood, \citet{NIEMEYER2014152} proposed a contextual classification method based on Conditional Random Field (CRF). A linear and a random forest models were compared when used for both the unary and the pairwise potentials.
By considering complex interactions between points, promising results were achieved, although this came at the cost of computation speed: 3.4 minutes for testing using an RF model, and 81 minutes using the linear model. 
The speed excludes the additional time needed to estimate the per-point, 131-dimensional feature vector prior to testing.
%The ISPRS 3D Semantic Labeling Benchmark was used, with results applied to building reconstruction. 
%Both a linear and  were compared for both the unary and the pairwise potentials. 
%During the training process in the first part, they randomly sampled 2000 point per-class for training, while testing was carried out using the remaining points. 
%We believe that randomly taking samples from test scene for training is not ideal which in turn will result in a biased prediction since training is done on parts of the test set. 
%Also, they had to use multiple complex contextual models to capture the interactions of individual points with their surroundings. 
 
%{\mxy Compared to our work, we don't explicitly use contextual model, instead we learn global features per-region during the training process using only simple input features that are readily available}. 

This contextual classification model was later extended to use a two-layer, hierarchical, high-order CRF, which incorporates spatial and semantic context \citep{NIEMEYER2016}. 
The first layer operates on the point level (utilizing higher-order cliques and geometric features \citep{weinnman2014}), to generate segments. 
%which operates on a point level and utilizes high-order cliques and a robust $P^{n}$ Potts model known as the point-layer. 
% \citep{weinnman2014}. 
%that is known to be time consuming. 
The second layer operates on the generated segments, and therefore incorporates a larger spatial scale. 
Used features include geometry and intensity-based descriptors, in addition to distance and orientation to road features \citep{Golovinskiy}). 
%, such as those derived from  normal-based features in addition to two road-related features namely the {\it distance to a road} and the {\it orientation with respect to the closest road} . 
By iteratively propagating context between layers, incorrect classifications can be revised at later stages; this resulted in good performance on the ISPRS 3D Semantic Labeling Contest. 
However, this method employed multiple algorithms, each designed separately, which would make it challenging to simultaneously optimize. Also, the use of computationally-intensive inference methods would limits the run-time performance.
%First, the method consists of multiple algorithms where each is designed separately. 
%For example, the feature estimation process, the probabilistic classifier of the unary potential, and the minimization of the CRF energy function are separate algorithms that are learned or trained individually. 
%Additionally, the use of complexinference methods such as graph-cut and Loopy Belief Propagation (LBP) at different stages, in addition to the time needed for estimating the features, increases the inference time during testing. 
In contrast to relying on multiple individually-trained components, an end-to-end learning mechanism is desired. 

%%===========================================================================================
\subsection{Indirect Methods}
\label{sec:indirect_methods}
Indirect methods -- which mostly rely on deep learning -- offer the potential to learn local and global features in a streamlined, end-to-end fashion \citep{deepvis}. 
Driven by the reintroduction and improvement of Convolutional Neural Networks (CNNs) \citep{LeCun1989b,resnet}, the availability of large-scale datasets \citep{imagenet}, and the affordability of high-performance compute resources such as graphics processing units (GPUs), deep learning has enjoyed unprecedented popularity in recent years. 
This success in computer vision domains such as image labeling \citep{Alex_NIPS2012}, object detection \citep{RCNN2014}, semantic segmentation \citep{Segnet_PAMI,FCN}, and target tracking \citep{DLT,SPT}, has generated an interest in applying these frameworks for 3D classification. 

However, the nonuniform and irregular nature of 3D-point clouds prevents a straightforward extension of 2D-CNNs, which were designed originally for imagery. Hence, initial deep learning approaches have focused on 3D computer-aided design (CAD) objects, and have relied on \emph{transforming} the 3D data into more tractable 2D images. 
For example, \citet{MVCNN} rendered multiple synthetic ``views'' by placing a virtual camera around the 3D object.
Rendered views were passed though replicas of the trained CNN, aggregated using a view-pooling layer, and then passed to another CNN to learn classification labels. 
%During training, the weights are initialized using VGG-M network \citep{vggm}. 
Several other methods use the multiview approach with various modifications to the rendered views. 
For example, \citet{GIFT} generated depth images as the 2D views, while other methods accumulated a unique signature from multiple view features, or projected the 3D information into 36 channels, modifying AlexNet \citep{Alex_NIPS2012} to handle such input. 
For further details, the reader is referred to \citep{shrec}. 

%=====================================================================================

Similar multiview approaches have also been applied to ground-based LiDAR point clouds.
For example, \citet{Boulch_3D} generated a mesh from the Semantic3D Large-scale Point Cloud Classification Benchmark \citep{hackel2017semantic3d}; this allowed for the generation of synthetic 2D views based on both RGB information and a 3-channel depth composite. 
A two-stream SegNet \citep{Segnet_PAMI} network was then fused with residual correction \citep{Audebert2016SemanticSO} to label corresponding pixels. 
2D labels were then back-projected to the point cloud to generate 3D semantic classification labels. 
Likewise, \citet{Luca} generated multiple overhead views, embedded with elevation and density features, to assist with road detection from LiDAR data. 
A Fully-Convolutional Network (FCN) \citep{FCN} was used for a single-scale binary semantic segmentation \{road, not-road\}, based on training from the KITTI dataset \citep{KITTI}. 

Despite their initial adoption, such multiview transformation approaches applied to point clouds lose information on the third spatial dimension through a projective rendering.
Simultaneously, they introduce interpolation artifacts and void locations.
Together, this complicates the process by unnecessarily rendering the data in 2D, in addition to forcing the network to ignore artificial regions caused by the voids.
While this is less consequential to binary classification,  multi-class problems require each point be assigned a separate class; this increases the complexity and may reduce the network's performance. %====================================================================================

In light of these limitations of multiview transformation methods, other authors have taken a volumetric approach to handle points clouds using deep learning. 
For example \citet{Bo_Li_iros} presented a method for vehicle detection in ground-based LiDAR point clouds. 
The input point cloud was voxelized, and then a fourth binary channel was appended, representing the \emph{binary occupancy}, \ie~ the presence or the absence of a point within each voxel. 
A 3D-FCN was then trained and evaluated to produce two maps representing the objectness and bounding box scores, using the KITTI dataset. 
Similarly, \citet{huang2016point} generated occupancy voxel grids based on LiDAR point cloud data, labeling each voxel according to the annotation of its center point. 
A 3D-CNN was then trained to label each voxel into one of seven classes; individual points were then labeled according to their parent voxel.  
Other authors have explored variations of voxelization methods including, binary occupancy grid, density grid, and a hit grid.  
%In a binary grid, each voxel is assumed to have a binary state, occupied or unoccupied. 
%In the density grid model, each voxel is assumed to have a continuous density corresponding to the probability that the voxel would block a sensor beam. 
%Finally, the hit grid only consider hits and ignore the difference between unknown and free space. 
In VoxNet, \citet{maturana2015voxnet} tested each voxelization model individually, to train 3D-CNNs with 32x32x32 grid inputs.
To handle multi-resolution inputs, they trained two separate networks each receiving an occupancy grid with different resolution

Parallel development of both multiview and volumetric CNNs has resulted in an empirical performance gap.
\citet{qi2016volumetric} suggested that results could collectively be improved by merging these two paradigms.
%Two hybrid volumetric CNN architectures were proposed. 
%The first introduced auxiliary learning tasks (\ie~ classifying object parts), which enabled deeperin turn helped to implicit learning deeper details of the 3D-shape. 
To address this, a hybrid volumentric CNN was proposed, which used long anisotropic kernels to project the 3D-volume into a 2D-representation.
Outputs were processed using an image-based CNN based on the Network In Network (NIN) architecture~\citep{NIN}. 
To combine the multiview approach with proposed volumetric methods, the 3D-object was rotated to generate different 3D-orientations. 
Each individual orientation was processed individually by the same network to generate 2D-representations, which were then pooled together and passed to the image-based CNN. 

Finally, \Citet{Yansong} took a different approach by combining image-like representations with conditional random field in the context of data fusion. Instead of directly operating on the LiDAR data, they interpolated the DSM map as a separate channel. 
Using the imagery and the LiDAR data, two separate probability maps were generated. 
A pre-trained FCN was used to estimate the first probability map using optical imagery. 
Then, by handcrafting another set of features from both the spectral and the DSM map, a logistic regression was applied to generate a second set of probability maps. 
At the end of this two-stream process, the two probability maps were combined using high-order CRF to label every pixel into one of six categories.
%In contrast, we show that instead of creating two streams to account for the additional depth information, we can extend our single 3D-method to work with 2D-semantic segmentation with different modalities by a simple modification to the data preparation process.

\subsection{Contribution}
\label{sec:contribution}
Although indirect methods introduced the application of deep learning for the semantic labeling task, they typically require a transformation of the input data, \ie~ to views or volumes, in order to meet the ingest requirements of conventional image-based networks. 
Unfortunately, these transformations introduce computational overhead, add model complexity, and discard potentially relevant information.
Likewise, direct methods have relied on the proliferation of various handcrafted features, in addition to contextual relationships, in order to meet increasing accuracy requirements with simple discriminative models. 
This added complexity has come at the cost of computational efficiency. 

Meanwhile, the generation of 3D point cloud data has increased rapidly in recent years due to the availability of high-resolution optical satellite/airborne imagery, and the explosion of modern stereo photogrammetry algorithms leveraging new computational resources. 
Such algorithms triangulate 3D-point coordinates directly from the optical imagery, and thus retain inherent spectral information; these  attributes should be considered in the development of a successful model.
In order to scale the semantic classification task to meet the demands of emerging data volumes -- potentially at sub-meter resolution and global in coverage -- an efficient, streamlined, and robust model that directly operates on 3D point clouds is needed.

It is in this context that we propose a simple, fully-convolutional network for direct semantic labeling of 3D point clouds with spectral information. 
Our proposed approach utilizes a modified version of PointNet~\citep{pointnet}, a deep network which operates directly on point clouds and so provides a flexible framework with large capacity and minimal overhead for efficient operation at scale. 
Moreover, it respects the permutation-invariance of input points, and therefore avoiding the need to transform the points to images or volumes. 
%It provides a unified architecture for applications ranging from object classification to part segmentation, to semantic labeling. 
%While such methods showed great progress, we believe that labeling 3D-point clouds is best done directly without the need to change the nature of the data itself, unstructured and unordered. 
%Also, aerial point clouds capture various different regions on Earth (\eg~ different building types, materials, terrains, etc.) which requires a framework that is flexible, has a large capacity, and generalizes easily given the differences between such regions. 
Inspired by the success of PointNet in applications such as object classification, part segmentation, and semantic labeling, we make the following contributions:
\begin{enumerate}
\item We extend PointNet to handle complex 3D data, obtained from overhead remote sensing platforms using a multi-scale approach.
Unlike CAD models, precisely-scanned 3D objects, or even indoor scenes, airborne point clouds exhibit unique characteristics, such as noise, occlusions, scene clutter, and terrain variation, which challenges the semantic classification task.
\item We present a deep learning algorithm with convolutional layers that consume unordered and unstructured point clouds directly, and therefore respects the pedigree of the input 3D data without modifying its representation and discarding information. 
%Unlike existing methods that rely on gridding and two stream process \citep{Luca,Yansong}, creating image-like representations \citep{MVCNN,GIFT}, or segment the data \citep{ramiya2014semantic}, our method doesn't modify the nature of the 3D-point cloud.
\item We eliminate the need for calculating costly handcrafted features, and achieve near state-of-the-art results with just the three spatial coordinates and three corresponding spectral values for each point. At the same time, the overhead of adding additional features to our model is minimal compared to adding new channels or dimensions in 2D
and volumetric cases.
\item We avoid the need to explicitly calculate contextual relationships (\eg~ by using CRF) and instead use a simple layer that can learn a per-block global features during training. 
\item Being fully convolutional, our network mitigates the issue of non-uniform point density, a common pitfall for stationary LiDAR platforms. 
%Hence, it is able to consume regions with different point densities.
\item We show that test time is on the order of seconds, compared to minutes for existing techniques relying on handcrafted features or contextual methods, and operating on the same dataset.
\item Finally, we show how the network can easily be applied to handle the issue of multimodal fusion in 2D-semantic segmentation, with a simple modification to the data preparation step. 
\end{enumerate}

%<><><><><><><><><><><><><><><><><><><><><><><><><><><><><><><><><><><><><><><><><><><><><><><><><>
\section{Methodology}
\label{Methodology}
%<><><><><><><><><><><><><><><><><><><><><><><><><><><><><><><><><><><><><><><><><><><><><><><><><>
%As reviewed in Section \ref{sec:related_work}, direct methods such as SVM or RF operate on each point independently, while CRF methods incorporate additional processing layers to account for contextual information. 
%Since such methods are well established in the literature, a large amount of time is devoted to the engineering of a reliable set of features separately from the classification algorithm. 
In this section we present a CNN-based deep learning method that is able to learn point-level and global features directly in an end-to-end fashion, rather than relying upon costly features or contextual processing layers. 
In Section \ref{sec:adaptation_of_cnn_to_point_clouds}, we describe how convolutional networks can be adapted to irregular point cloud data.
%We detail the specific layers used in Section \ref{sec:convolutional_layers}, and introduce an activation function for learning more complex features. 
Section \ref{sec:batch_normalization} describes how batch normalization is used to precondition the outputs of the activation functions. 
Section \ref{sec:contextual_feature_learning} describes a pooling layer that is used to learn contextual features.
Finally, Section \ref{sec:inference} details the inference of semantic classification labels from the learned local and global features.
%The following subsections describe the details of our deep learning framework.
%<><><><><><><><><><><><><><><><><><><><><><><><><><><><><><><><><><><><><><><><><><><><><><><><><>
\subsection{Adaptation of CNN to Point Clouds}
\label{sec:adaptation_of_cnn_to_point_clouds}
%<><><><><><><><><><><><><><><><><><><><><><><><><><><><><><><><><><><><><><><><><><><><><><><><><>
CNN architectures consist of multiple, layered convolution operations, wherein at each layer, the set of filter weights is learned based on training data for a specific task. 
Recall that for a single 2D-convolution (Equation \ref{eq:1})  a filter $(h)$ in layer $(\ell)$ and channel $(d)$ ``slides'' across the domain of the input signal, $x[u,v]$, accumulating and redistributing this signal into the output, $f^{(\ell,d)}[m,n]$.
 
%The size of the filter -- also called the receptive field or \emph{weights} -- depends on the underlying task and is usually determined empirically.
\begin{equation}
\label{eq:1}
f^{(\ell,d)}[m,n] = \sum_{u=-\infty}^{\infty} \sum_{v=-\infty}^{\infty} x[u,v] h^{(\ell,d)}[m-u,n-v]
\end{equation}
This sliding process can be thought of as replicating the filter at every spatial location.
Replicating the filter in this way allows for the extraction of features regardless of their position, and enables a linear system to be shift-invariant. 
Additionally, sharing the same set of weights at multiple locations increases the learning efficiency by reducing the number of parameters in the model. 
Based on the nature of the convolution, CNNs typically require highly regular data, \eg~ images, which are organized based on a 2D-grid. 
Also, note that such a convolution (applied to gridded data, \eg~ images), is not invariant to permutations of the input members \ie~ pixels. 
In other words, the spatial distribution of pixels within a filter window is important to capture local features such as edges. Therefore, reordering the input points will result in meaningless output. 

This introduces challenges for the application of CNNs to classify irregular, unstructured 3D point cloud data. 
Given an input set of $N$ 3D-points $X$ = \{$x_{1},x_{2},x_{3},...,x_{N}$\}, where every point represents a row in a 2D-array, the goal of point cloud classification is to assign every point $x_{i}$ an object-level label from a set of predefined labels $Y$ = \{$y_{1},y_{2},y_{3},...,y_{C}$\}, where $C$ is the total number of classes. 
Since point clouds are not defined on regular grids, and convolutional layers require regular inputs, a modification to either the input or the network architecture is needed. 
In order to directly operate on point clouds and avoid transforming the data to a different representation (see Section \ref{sec:indirect_methods}), we follow~\citep{pointnet} by adapting convolutional operations to point clouds. 

\begin{figure*}[th]
\begin{center}
\includegraphics[width=16cm]{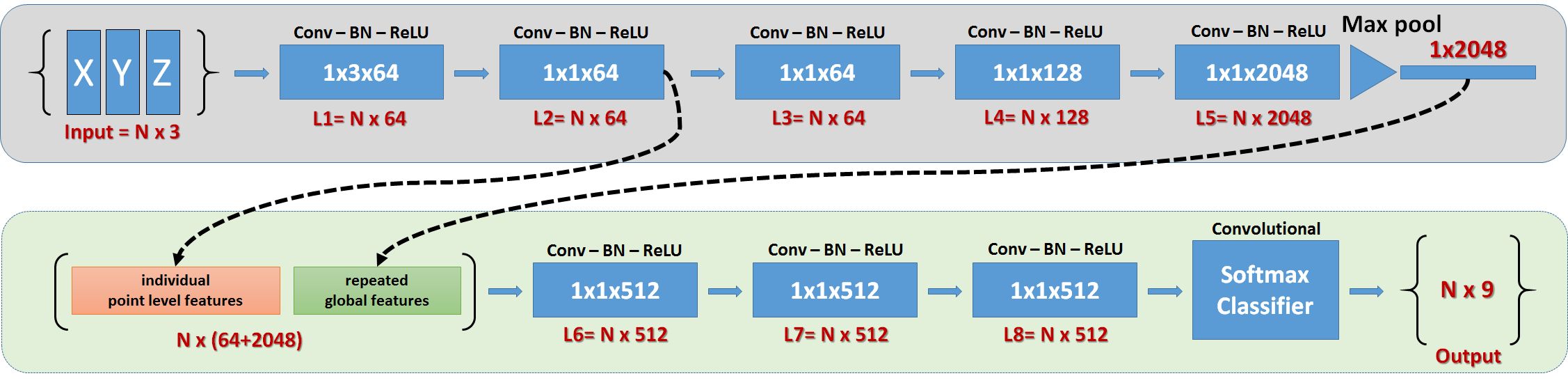}
\end{center}
\caption{The basic semantic labeling network takes as input an Nx3 set of points (a point ``cloud''), and, in the first stage, passes it through a series of convolutional layers to learn local and global features. In the second stage, concatenated features are passed though $(1\times1)$ convolutional layers and then to a softmax classifier to perform semantic classification. Text in white indicates the filter size, while text in red indicates the layer's output shape.}
\label{fig:algorithm}
\end{figure*}
%\subsection{Convolutional Layers}
%\label{sec:convolutional_layers}

The complete architecture of our network is shown in Figure \ref{fig:algorithm}. 
The input to the network is an $N\times M$ array of unordered data points, where $N$ is the number of points, and $M$ is the number of features for each point, \ie~ spatial coordinates and/or spectral information. 
Shown in Figure \ref{fig:algorithm} is the simple case where the input data is the raw point cloud $X$, defined by its spatial coordinates (x,y,z) as columns of the array. The input could optionally be expanded to include any other features, such as spectral information.
The first layer of the network applies a 1D-convolution -- with a filter width equal to the width of the input vector -- across the columns to capture the interactions between coordinates for each row (data point).
The output of this layer is a single column where each value corresponds to an individual 3D-point within the set, \ie~ an $(N\times M)$ input array is transformed to an $(N\times1)$ output array.  
This layer operates on each point independently, an advantage which allows for the incorporation of point-wise operations such as scaling, rotation, translation, etc. 
Since such operations are differentiable, they can be included within the network as a layer and trained in an end-to-end fashion. 
This concept, first introduced in~\citep{spatial_trans_net}, allows the network to automatically align the data into a canonical space and therefore makes it invariant to geometric transformations. 
%In our network, we didn't include a transformation layer, but we handle rotations by augmentation as described in Section \ref{sec:preprocessing}.  

The subsequent convolutional layers perform feature transformation -- such as dimensionality reduction or feature expansion -- using $(1\times1)$ convolutions.We avoid fully-connected layers due to their expensive computational cost. 
Since the convolution process is a linear operation that performs a weighted sum of its inputs, a non-linear operation -- known as the activation function -- is needed in order to derive and learn more complex features. 
Three activation functions are frequently used in the literature: $sigmoid$, hyperbolic tangent ($tanh$), and rectified linear unit ($ReLU$). 
Sigmoid activation, \(\sigma(x)=1/(1+e^{-x})\), reduces the input values to the range of [0,1]; this can be thought of as assigning a likelihood value to the input. 
Similarly, $tanh$ activation, \(\tanh(x) = 2 \sigma(2x) -1\), maps the input values to the range of [-1,1]; this has the added advantage that the output is zero centered. 
Finally, $ReLU$ activation, $f(x) = \max(0, x)$, applies a simple ramp function. 
It reduces the likelihood of vanishing gradient, greatly accelerates the convergence rate \citep{Alex_NIPS2012}, and involves simpler mathematical operations; therefore it is the most common activation function used in deep convolutional networks.  
We implement the $ReLU$ function in our network as follows:
\begin{equation}
\label{eq:2}
f^{(1)} = \max(0,\langle h, x_{i} \rangle +b)
\end{equation}
where the convolution in Equation \ref{eq:2} is represented now by the dot product $\langle \cdot \rangle$, $b$ is the bias, and $f^{(1)}$ is the first layer's output. 
%<><><><><><><><><><><><><><><><><><><><><><><><><><><><><><><><><><><><><><><><><><><><><><><><><>
\subsection{Batch Normalization}
\label{sec:batch_normalization}
%<><><><><><><><><><><><><><><><><><><><><><><><><><><><><><><><><><><><><><><><><><><><><><><><><>
Although the $ReLU$ activation function has many advantages, it does not enforce a zero-centered distribution of activation values, which is a key factor to improve the gradient flow. 
%However, it is clear from Equation \ref{eq:2} that there is no control on the distribution of the activation function output, except that it is always positive. 
One way to adjust this distribution is to change the weight initialization mechanism.  
\Citet{pmlr-glorot10a} and \citet{He_Normal} showed that good initialization is the key factor for a successful, convergent network, however, control over the distribution of activations was handled indirectly.
For improved control, \citet{BatchNorm} introduced Batch Normalization (BN) that directly operates on the activation values. 
An empirical mean and variance of the output (after the convolutional layer and before the non-linearity) are computed during training; these are then used to standardize the output values,  \ie 
%However, if a network is initialized with all weights being zero, all the hidden layers will end up having the same value during the forward and the backward passes, leading to the undesired symmetric ways phenomenon. 
%This due to the activation functions having a spiked distribution \eg~~\(\mathcal{N}(0,0)\) that only covers a single value. 
%To avoid this issue, weights can be initialized by randomly sampling a Gaussian distribution with a small variance. 
%However, although this works well for small networks, it can lead to non-homogeneous distributions of activation values in deep networks. 
%{\mxy This can be proven as follows. Let $\psi(x)$ be a 10-layers fully connected neural network operating on the input $x$ with the weights of each layer being initialized using $h\sim N(0,10^{-4})$. Let each layer has 500 nodes with tanh non-linearity. Plotting the histogram of the activation functions in each layer show that the distribution becomes very narrow when when we go deeper in the network as shown in {\mxy Figure2}}. 
%To better initialize large networks, \citet{pmlr-glorot10a} suggested instead using the variance of \(1/fan_{in}\), where $fan_{in}$ is the number of incoming connections, in order to XXX. 
%While these efforts enabled new methods which are not reliant upon per-layer pretraining, its mathematical derivation assumed linear activations. 
%Subsequent work by ~\citet{He_Normal} extended this approach to nonlinear $ReLU$ activations. 
%However, in both of these methods, control over the distribution of the activations is handled indirectly. 
\begin{align}
\label{eq:batchnorm}
\begin{split}
\hat{s} &= \frac{{s} -E[s]}{ \sqrt{Var[s]} }
\\
 z &= \gamma\cdot\hat{s}+\beta
\end{split}
\end{align}
where $s=\langle h{,}x_{i}\rangle$ is the output after the convolutional layer, and $\gamma$ and $\beta$ are the scale and shift parameters learned during training. 
Setting $\gamma= \sqrt{Var[s]}$, and $\beta=E[s]$ allows the network to recover the identity mapping.
BN improves flow through the network, since all values are standardized, and simultaneously reduces the strong dependence on initialization. 
Also, since it allows for homogeneous distributions throughout the network, it enables higher learning rates, and acts as a form of regularization. 

Incorporating these advantages, we initialize weights using the method of~\citet{pmlr-glorot10a} and insert BN layers after every convolutional layer, as shown in Figure \ref{fig:algorithm}. 
Note that the BN layer functions differently during testing and training. 
During testing, the mean and the variance are not computed. 
Instead, a single fixed value for the mean and the variance -- found empirically during training using a running average -- is used during testing. 
After integrating BN, the activation function (Equation \ref{eq:2}) can now be written as follows:
\begin{equation}
\label{eq:3}
f = \max(0,BN(h^Tx+b))
\end{equation}
%The output of Equation \ref{eq:3} is a single column where each value corresponds to an individual 3D-point within the set \ie~ going from $(N\times3)$ input array to $(N\times1)$ output array. 
Evaluating Equation \ref{eq:3} multiple times for different values of $h$ allows each layer to capture various aspects of its input. 
This results in an output array of $(N\times K)$ dimensions, where $K$ is the total number filters used. 
The size of $K$ per-layer is shown in Figure \ref{fig:algorithm} (white text). Following the output with a series of convolutional layers, as shown in the upper part of Figure \ref{fig:algorithm}, can be defined mathematically as a sequence of nested operations as follows:
\begin{equation}
\label{eq:4}
f^{(\ell)} = f^{(\ell-1)}(f^{(\ell-2)}(...f^{(1)}(x)))
\end{equation} 
where $f^{\ell}$ is defined as in Equation \ref{eq:3}, and $\ell$ is the layer index. 
%<><><><><><><><><><><><><><><><><><><><><><><><><><><><><><><><><><><><><><><><><><><><><><><><><>
\subsection{Contextual Feature Learning}
\label{sec:contextual_feature_learning}
%<><><><><><><><><><><><><><><><><><><><><><><><><><><><><><><><><><><><><><><><><><><><><><><><><>
During training, we desire a network that can learn both local features ( at the individual point-level) and global features, which provide additional contextual information to support the classification stage. 
Here, we describe how global features can be extracted using a pooling layer, which simultaneously provides permutation-invariance, \ie~ the order of the input points does not affect classification results. 
Contrary to 2D-images, point clouds are unstructured and unordered, and so an appropriate network should respect this original pedigree. 

Three options are available: sorting the data as a preprocessing step, using a Recurrent Neural Network (RNN), or incorporating a permutation-agnostic function that aggregates the information from all points. 
In the first case, the optimal sorting rules are not obvious.
In the second case, the point cloud must be considered as a sequential signal, thus requiring costly augmentation of the input with all possible permutations. 
While some architectures such as long short-term memory (LSTM) and gated recurrent unit (GRU) neural networks can deal with relatively long sequences, it becomes difficult to scale to millions of steps, which is a common size in point clouds. 

Given these considerations, we incorporate a permutation-agnostic function as an additional layer.
Specifically, pooling layers, commonly used to downsample 2D-images, work perfectly for this purpose. 
Instead of downsampling the point set, we pool values across all points to form a single global signature that represents the input point cloud. 
Such signature could be used directly to label the whole set. 
However, recall that for the task of semantic labeling, a label for each 3D-point is desired. 
Therefore, both local \emph{and} global features are needed to describe the point  and capture contextual information within the set. 

In this network, local features are obtained from the second convolutional layer $f^{(2)}$ with an output shape of $(N\times64)$, \ie~ this represents each 3D-point with a 64D-vector. 
On the other hand, a global signature for the point set is derived from the output of the fifth convolutional layer with dimensions of $(N\times2048)$.
This serves as input to the global feature extraction function, specifically, a {\it max-pooling} layer, which aggregates the features across all points and produce a signature with a shape of $(1\times2048)$ as follows:
\begin{equation}
\label{eq:5}
g = \max_{row}(f^{(5)})
\end{equation}
where $g$ is the global feature vector, $f^{(5)}$ is the output at the $5^{th}$ layer, and $row$ indicates that the aggregation is applied vertically across the rows, \ie~ points. 
%<><><><><><><><><><><><><><><><><><><><><><><><><><><><><><><><><><><><><><><><><><><><><><><><><>
\subsection{Inference}
\label{sec:inference}
%<><><><><><><><><><><><><><><><><><><><><><><><><><><><><><><><><><><><><><><><><><><><><><><><><>
The global feature vector is concatenated with the point level features, yielding a  per-point vector that contains both local and contextual information necessary for point-wise labeling. 
This concatenated feature vector is then passed to a series of feature transformation layers and finally to a softmax classifier. 
We use the cross-entropy cost function to train the network. 
Cross-entropy is a special case of the general Kullback-Leibler (KL)-divergence $D_{\text{KL}}$, which measures how the ground-truth probability distribution $p$ diverges from the output probability distribution $q$, \ie~:
\begin{equation}
\label{eq:6}
\begin{array}{rl}
D_{\text{KL}}(p||q)
    & = \sum^{}_{i} p(x_{i})\cdot(\log p(x_{i}) - \log q(x_{i})) \\
    & = -\sum^{}_{i} p(x_{i}) \log q(x_{i}) -\sum^{}_{i} p(x_{i}) \log \frac{1}{p(x_{i})}  \\
    & = H(p,q) - H(p)
\end{array}
\end{equation}
If the discrete distribution $p$ is zero everywhere except a single location with maximum probability, the expression is reduced to the cross-entropy $H(p,q)$. 
In our case, $p$ is the ground truth distribution represented by one-hot vectors encoding the label per-point, while $q$ is the output of the softmax layer, representing the normalized class probabilities.
Here, each individual class probability is calculated as follows:
\begin{equation} \label{eq:7}
P(y_i \mid f^{(\ell)}(x_i); W) = \frac{e^{f_{y_i}}}{\sum_c e^{f_c} }
\end{equation}
where $f^{(\ell)}$ is the last convolutional layer and the input to the softmax layer, $y_i$ is the current correct label for the input $x_i$, $W$ is the weight matrix for the softmax classifier, and $f_{y_i}$ is the unnormalized log probability of the output node indexed by $y_i$, and $c$ is the class (output) index.
%<><><><><><><><><><><><><><><><><><><><><><><><><><><><><><><><><><><><><><><><><><><><><><><><><>
\section{Evaluation}
\label{sec:evaluation}
%<><><><><><><><><><><><><><><><><><><><><><><><><><><><><><><><><><><><><><><><><><><><><><><><><>
\begin{figure*}[th]
\begin{center}
\includegraphics[width=16cm]{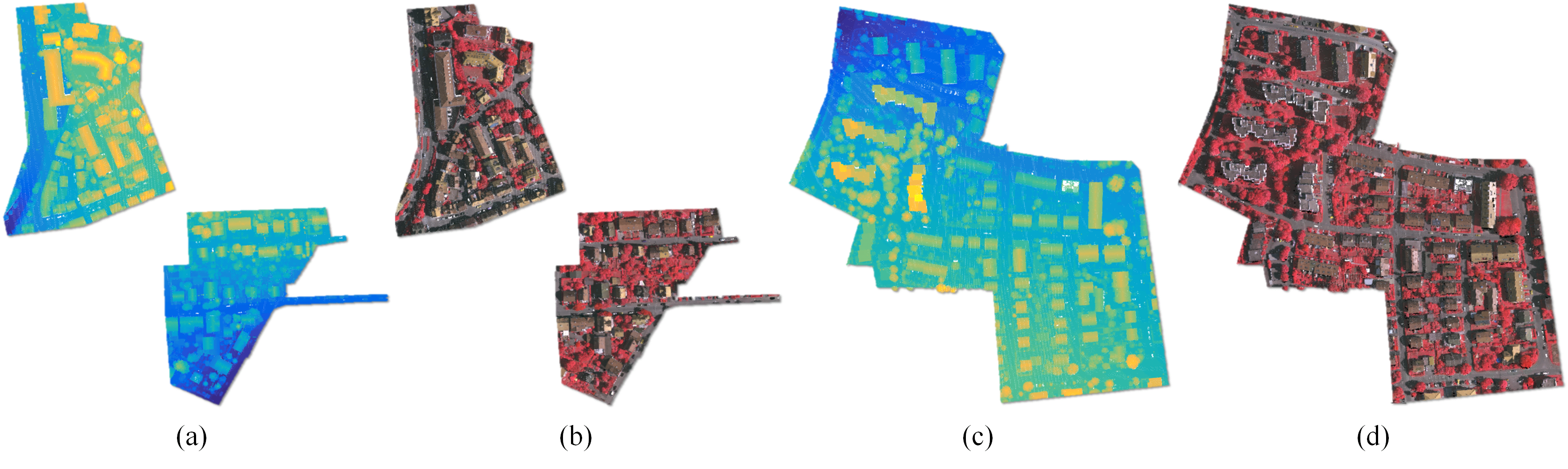}
\end{center}
\caption{From left to right: point cloud (a) color-coded by height, and (b) by spectral information (IR,R,G) for the test area; and point cloud (c) color-coded by height, and (d) by spectral information (IR,R,G) for the training data.}
\label{fig:dataset_color}
\end{figure*}
This section describes the methodology used to evaluate the performance of our approach. 
In Section \ref{sec:dataset} we describe the datasets used. 
In Section \ref{sec:preprocessing} we describe preprocessing steps. 
In Section \ref{sec:training_parameters} we outline training parameters. 
In Section \ref{sec:classification_results} we present our results on 3D-point clouds along with various experiments showing qualitative performance of our method when applied to different testing cases. 
In Section \ref{sec:effect_of_individual_features} we analyze the effects of the input feature selection. 
Finally, in Section \ref{sec:extension_to_2D} we show the extension of our network to handle the fusion of LiDAR and spectral data in 2D-semantic segmentation tasks.
%<><><><><><><><><><><><><><><><><><><><><><><><><><><><><><><><><><><><><><><><><><><><><><><><><>
\subsection{Dataset}
\label{sec:dataset}
%<><><><><><><><><><><><><><><><><><><><><><><><><><><><><><><><><><><><><><><><><><><><><><><><><>
%Our first contribution is adapting the vanilla version of PointNet network and apply it to complex scenes such as aerial point clouds instead of CAD models. In this paper we use the airborne laser scanning data from Vaihingen, Germany used for the ISPRS 3D labeling challenge as part of the urban classification and 3D reconstruction benchmark In total 9 classes have been defined including {\it Powerline}, {\it Low vegetation}, {\it Impervious surfaces}, {\it Cars}, {\it Fence/Hedge}, {\it Roof}, {\it Facade}, {\it Shrub}, and {\it Tree}. The area is divided into two parts, training and testing. Each part comes with a text file that includes the (x, y, z) coordinates, reflectance and return count information acquired using Leica ALS50 system at a mean height of 500m above ground. The point density in both parts is approximately 8 points/m2. The test area is within the center of Vaihingen city which is characterized by dense, complex buildings. The training area on the other hand, is mostly residential with detached houses and high rise buildings. While the data doesn’t include spectral information, the corresponding georeferenced IR-R-G imagery from the 2D semantic labeling contest is used to extract spectral data. Our goal in this paper is to show that without handcrafting 3D features or explicitly using contextual methods, our network can achieve near state-of-the-art results. Figure2 shows the original point clouds with and without spectral information.
For this paper, we use data provided by both the ISPRS 2D\footnote{\url{https://goo.gl/mdvbwM}} and 3D-Semantic Labeling Contest, as part of the urban classification and 3D-reconstruction benchmark.
Both airborne LiDAR data and corresponding georeferenced IR-R-G imagery are provided from Vaihingen, Germany (Figure \ref{fig:dataset_color}).
For the 3D contest, 9 classes have been defined, including {\it Powerline}, {\it Low vegetation}, {\it Impervious surfaces}, {\it Cars}, {\it Fence/Hedge}, {\it Roof}, {\it Facade}, {\it Shrub}, and {\it Tree}. 
The area is subdivided into two regions, for training and testing. 
Each region includes a text file that contains the LiDAR-derived (x,y,z) coordinates, backscattered intensity, and return count information, acquired using a Leica ALS50 system at a mean height of 500m above ground. 
The point density is approximately 8 points/m$^{2}$. 
The test area is within the center of Vaihingen city, and is characterized by dense, complex buildings. 
The training area, on the other hand, is mostly residential, with detached houses and high rise buildings. 

%Our goal in this paper is to show that without handcrafting 3D-features or explicitly using contextual methods, our network can achieve near state-of-the-art results while operating on unordered and unstructured point cloud. 
%Figure \ref{fig:dataset_color} shows the original point clouds with and without spectral information.
%<><><><><><><><><><><><><><><><><><><><><><><><><><><><><><><><><><><><><><><><><><><><><><><><><>
\subsection{Preprocessing}
\label{sec:preprocessing}
%<><><><><><><><><><><><><><><><><><><><><><><><><><><><><><><><><><><><><><><><><><><><><><><><><>
Two preprocessing methods were employed to obtain our desired input. 
First, spectral information was attributed to each (x,y,z) triplet in the point cloud by applying a bilinear interpolation using the georeferenced IR-R-G imagery as shown in Figure \ref{fig:dataset_color}. 
Note that in the case of stereo-derived point clouds from optical imagery, this spectral information is inherently available, and would not need to be obtained separately.  
Next, we normalize the z values in the point cloud by subtracting a Digital Terrain Model (DTM), generated using LAStools\footnote{\url{http://www.lastools.org/}}, in order to obtain height-above-ground. Then, to train our deep learning method from scratch, a sufficiently large amount of labeled data are required; however, a single and small training scene is provided. 
We solve this issue by subdividing our training and testing regions into smaller 3D-blocks. 
Such blocks are allowed to overlap (unlike PointNet), thus increasing the quantity of data available, and robustness by allowing overlapped points to be part of different blocks. 
Each point within the block is represented by a 9D-vector, containing the per-block centered coordinates (X,Y,Z), spectral data (IR,R,G), and normalized coordinates (x,y,z) to the full extent of the scene. 
Note that since our method is fully convolutional, the number of points per-block can vary during training and testing.
This contribution resolves the typical challenge of working with point clouds of varying density. 
While we test using different densities, we sample fixed number of points per-block during training for debugging and batch training purposes. 

To sample points from each block, we randomly choose 4096 points during training without replacement.
If the number of points per-block is lower than the desired number of samples, random points within the block are repeated. 
However, if the number of points per-block is lower than 10 points, the block is ignored. 
To learn objects with different scales, \eg~ building vs. car, one can train separate networks, with each network trained using a down-sampled version of the point cloud, as in \citet{maturana2015voxnet}. 
However, this is not practical, as it introduces an additional and unnecessary computational overhead. 
Instead, current deep learning approaches -- \eg~ a single network with high capacity -- should be able handle multi-scale objects in an end-to-end fashion given appropriate inputs. 
To handle different resolutions, we generate blocks with different sizes and train our network using all scales simultaneously. % during training and testing. 
Blocks of size 2m$\times$2m, 5m$\times$5m, and 10m$\times$10m work well in our case given the scale of the features of interest \eg~ cars and roofs. 
Our final result during testing is the average of all three scales.
While splitting the data into blocks with different sizes increases the number of training samples, robustness to noise and orientation could be further improved by augmenting the training data with modified versions of the original training data. 
We augment the training data by randomly rotating points around the z-axis (\ie~ to adjust their geographic orientation), and jittering the coordinates.  
Jitter is added by applying an additive noise, sampled from a zero-mean normal distribution with $\sigma=0.08$m for the $x,y$ coordinates, and $\sigma=0.04$ for the $z$ coordinates. 
Next we clip the values to a maximum horizontal and vertical jitter of 30cm and 15cm respectively. 
The values were chosen empirically to add sufficient noise while preserving the relative differences between various objects.
%This allows for, nominally, a maximum horizontal and vertical jitter of 30cm and 15cm respectively. 
Rotation and jitter are applied before splitting the data into blocks. However, we also apply jitter during the per-block sampling process, in order to avoid duplicating points. % which also can be viewed as generating new points within the block. 
%{\mxy fig.. shows the original data and the augmented versions with close up on the jitter}
%<><><><><><><><><><><><><><><><><><><><><><><><><><><><><><><><><><><><><><><><><><><><><><><><><>
\subsection{Training Parameters}
\label{sec:training_parameters}
%<><><><><><><><><><><><><><><><><><><><><><><><><><><><><><><><><><><><><><><><><><><><><><><><><>
To asses and monitor the performance of our model during training, a validation set is needed.
Instead of taking samples from the training set directly, we desire a validation set that is as different as possible to the training data.
We address this by splitting the original training data before augmentation into a new training and validation subsets using stratified splits to preserve the distribution of classes in both sets. 
Class-distributions within each set are then balanced by repeating under-represented classes until they match the class with the highest number of members, resulting in a uniform distribution of classes. 
We then augment the new training data by applying the jitter and rotations as described in Section~\ref{sec:preprocessing}. 
%We discard the new training set and train using the augmented data only. 
%While we could also add spectral noise, this step was not included in the current version. 

After splitting both the training and validation sets into blocks, we train using the augmented data only as input to the network shown in Figure~\ref{fig:algorithm}. 
We use the Adam optimizer~\citep{Kingma2014AdamAM} with an initial learning rate 0.001, a momentum of 0.9 and a batch size 32. 
The learning rate ($lr$) is iteratively reduced based on the current number of epochs, according to:
\begin{equation}
\label{eq:lr}
lr_{new} = lr_{initial}\times(1.0-\frac{epoch_{current}}{epoch_{total}})
\end{equation}
%<><><><><><><><><><><><><><><><><><><><><><><><><><><><><><><><><><><><><><><><><><><><><><><><><>
\begin{figure}[t]
\begin{center}
\includegraphics[width=8.1cm]{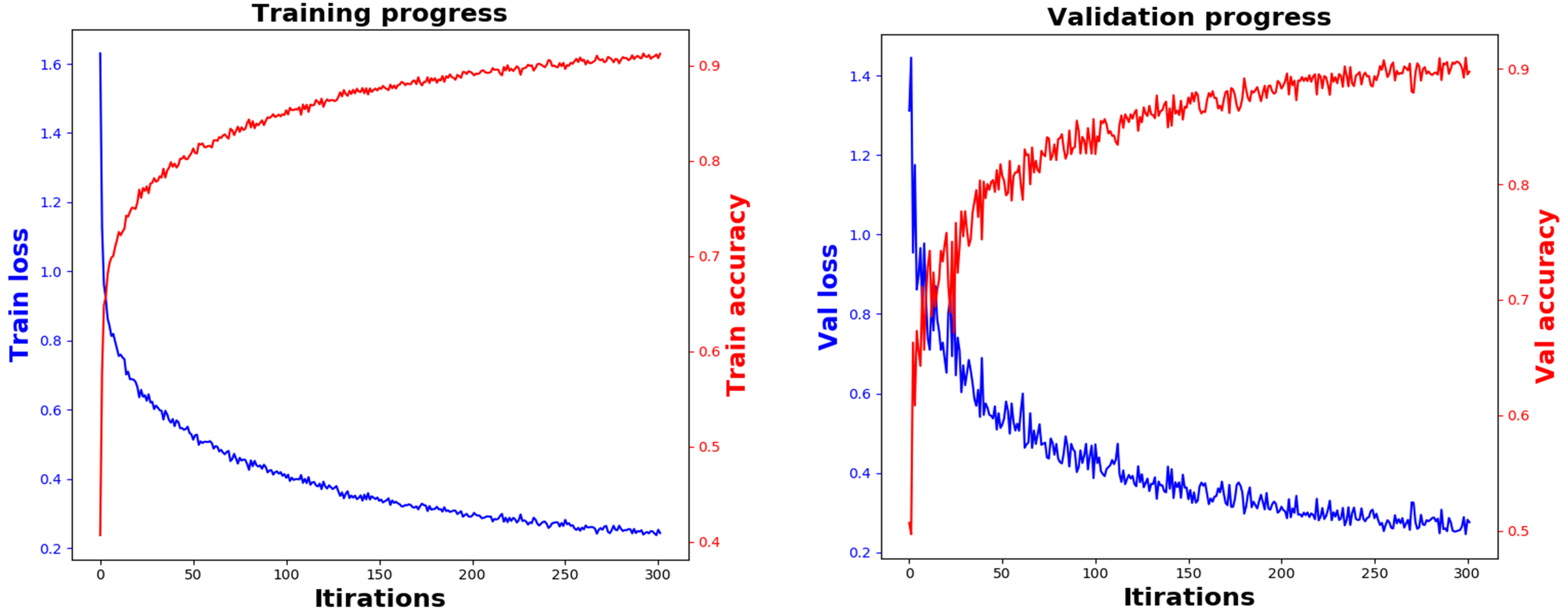}
\end{center}
\caption{The loss and overall accuracy progress for training (left) and validation (right). Every 10 iterations correspond to a single epoch.}
\label{fig:progress}
\end{figure}
This proceeds for a total of 30 epochs, \ie~ $epoch_{total} = 30$. 
We monitor the progress of the validation loss and save the weights if the loss improves. 
If the loss does not improve after 3 epochs, training is terminated and the weights with the best validation loss are used for testing. 
Training our network takes around 12 to 18 hours to converge using a Tesla p40 GPU and Keras~\citep{keras} with the Tensorflow backend.
The feed forward time during testing is 3.7s for the full scene ($\sim$ 412k points).  
Figure \ref{fig:progress} shows the loss and overall accuracy progress during training and validation.
%<><><><><><><><><><><><><><><><><><><><><><><><><><><><><><><><><><><><><><><><><><><><><><><><><>
\subsection{Classification Results}
\label{sec:classification_results}
\begin{figure}[t]
\begin{center}
\includegraphics[width=8cm]{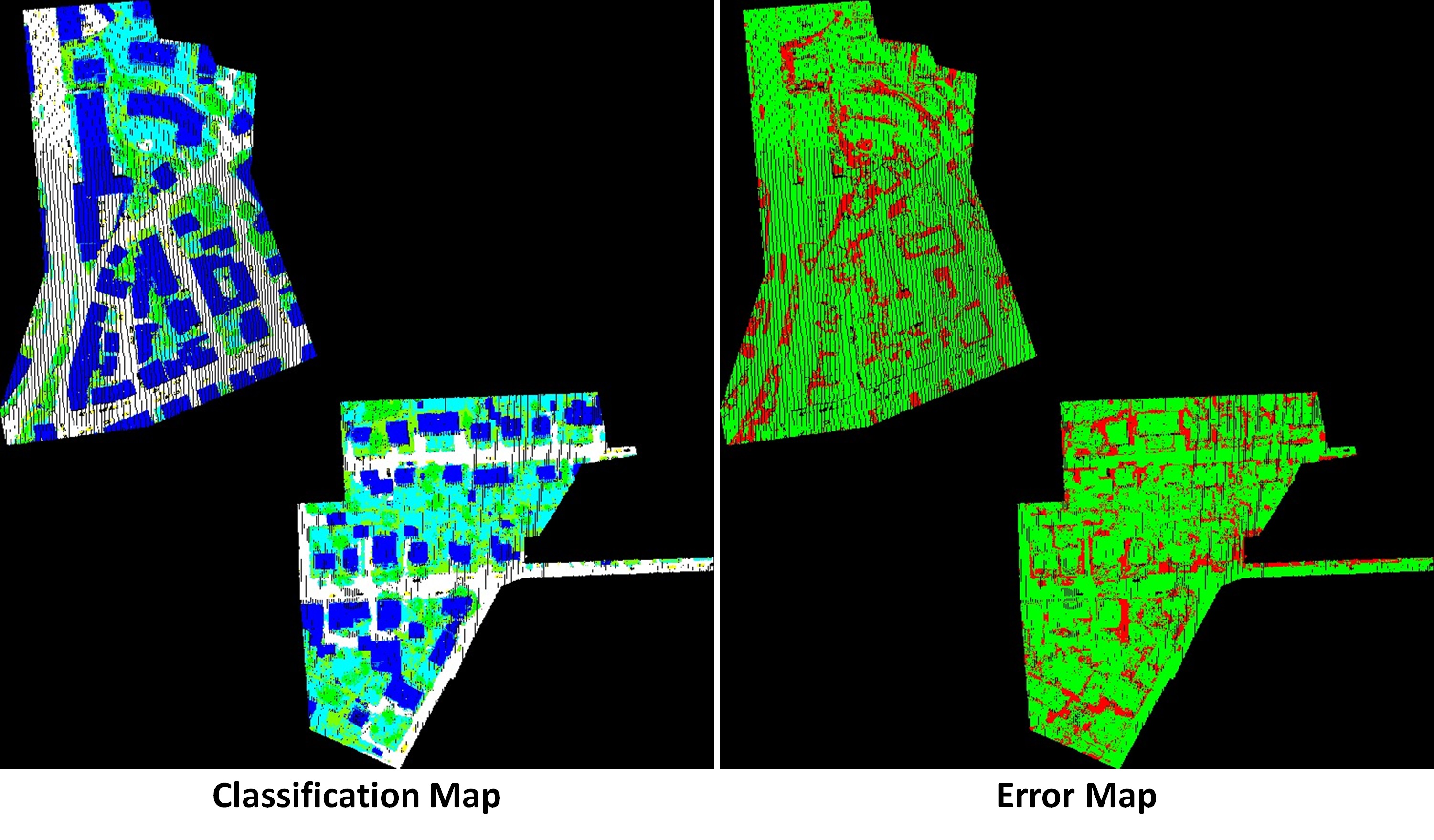}
\end{center}
\caption{The left image shows the classification map, while the right image shows the corresponing error map. The results were provided by the contest organizers.}
\label{fig:results&rg}
\end{figure}

%<><><><><><><><><><><><><><><><><><><><><><><><><><><><><><><><><><><><><><><><><><><><><><><><><>
\begin{table*}[t]
\centering
\caption{Confusion matrix showing the per-class accuracy using our deep learning framework. The overall accuracy (not shown) is \textbf{81.6}$\%$}
\label{tab:confusion}
\begin{tabular}{c|ccccccccc}
                \textbf{Classes} & \textbf{power} & \textbf{low\_veg} & \textbf{imp\_surf} & \textbf{car}   & \textbf{fence\_hedge} & \textbf{roof}  & \textbf{fac}   & \textbf{shrub} & \textbf{tree}  \\ \cline{1-10} 
\multicolumn{1}{c|}{\textbf{power}}                 & \textbf{29.8} & 00.0             & 00.2              & 00.0          & 00.0                 & 54.2          & 00.7          & 00.2          & 15.0 \\
\multicolumn{1}{c|}{\textbf{low\_veg}}              & 00.0          & \textbf{69.8}    & 10.5              & 00.5          & 00.2                 & 00.5          & 00.6          & 16.1          & 01.9 \\
\multicolumn{1}{c|}{\textbf{imp\_surf}}             & 00.0          & 05.2             & \textbf{93.6}     & 00.2          & 00.0                 & 00.2          & 00.1          & 00.7          & 00.0 \\
\multicolumn{1}{c|}{\textbf{car}}                   & 00.0          & 05.0             & 01.2              & \textbf{77.0} & 00.2                 & 02.6          & 00.8          & 12.9          & 00.3 \\
\multicolumn{1}{c|}{\textbf{fence\_hedge}}          & 00.0          & 05.9             & 01.4              & 01.7          & \textbf{10.4}        & 01.5          & 00.6          & 68.5          & 10.0 \\
\multicolumn{1}{c|}{\textbf{roof}}                  & 00.1          & 00.5             & 00.4              & 00.0          & 00.0                 & \textbf{92.9} & 02.8          & 02.3          & 00.9 \\
\multicolumn{1}{c|}{\textbf{fac}}                   & 00.2          & 04.3             & 00.8              & 00.9          & 00.1                 & 23.3          & \textbf{47.4} & 19.9          & 03.1 \\
\multicolumn{1}{c|}{\textbf{shrub}}                 & 00.0          & 07.9             & 00.5              & 01.0          & 00.5                 & 02.6          & 02.0          & \textbf{73.4} & 12.0 \\
\multicolumn{1}{c|}{\textbf{tree}}                  & 00.0          & 00.8             & 00.0              & 00.2          & 00.1                 & 01.2          & 01.3          & 17.1          & \textbf{79.3} \\ \hline
\multicolumn{1}{c|}{\textbf{Precision/Correctness}} & 50.4          & 88.0             & 89.6              & 70.1          & 66.5                 & 95.2          & 51.4          & 33.4          & 86.0 \\
\multicolumn{1}{c|}{\textbf{Recall/Completeness}}   & 29.8          & 69.8             & 93.6              & 77.0          & 10.4                 & 92.9          & 47.4          & 73.4          & 79.3 \\
\multicolumn{1}{c|}{\textbf{F1 Score}}              & 37.5          & 77.9             & 91.5              & 73.4          & 18.0                 & 94.0          & 49.3          & 45.9          & 82.5 \\ \hline
\end{tabular}
\end{table*}
Classification results are based on the ISPRS 3D Semantic Labeling Contest.
During testing, we split the data into blocks similar to the training stage, in order to recover objects at different scales. 
The block sizes, overlap, and number of points per-block are reported in Table \ref{tab:blocks}. %., in addition to the computation time needed for this step. 
%Note that the time needed to split the data into blocks could be easily reduced with parallel computation in future work. 
We forward pass the data to output a 9D-vector per-point indicating the probability of each point belonging to each of the nine categories. 
Since we use a fixed number of points for each block size, some points from the original data may remain unclassified due to sampling, while others may be duplicated due to repetition. 
Therefore, we interpolate the output results to classify the original set of points. 
We use nearest neighbor interpolation on each class probability separately, to generate a total of nine class probability maps. % as shown in fig{\mxy class maps prob}. 
A label corresponding to the index of the highest probability from the nine maps is assigned to each point, indicating the classification label.
\begin{table}[t]
\centering
\caption{The block sizes and the corresponding overlap and number of points during testing.}
\label{tab:blocks}
\begin{tabular}{c|c|c}
\textbf{Size} & \textbf{Overlap} & \textbf{\# Points} \\ \hline %& \textbf{Time(min)}  
2m$\times$2m            & 1m                   & 1024 \\        %& asd                
5m$\times$5m            & 2m                   & 3072 \\        %& asd                
10m$\times$10m           & 2m                   & 4096\\ \hline %& asd             
\end{tabular}
\end{table}
Quantitative performance metrics are based on the ISPRS contest: Per-class accuracy, precision/correctness, recall/completeness, and F1-score, in addition to the overall accuracy. 
Although it is possible to submit results while excluding some classes, we chose to evaluate our method on all available classes. 
The confusion matrix in Table~\ref{tab:confusion} shows resulting per-class accuracies.
Figure \ref{fig:results&rg} shows the corresponding classification map, along with the error map provided by the contest organizers. 
As shown in Table\ref{tab:confusion}, the proposed method performs well on {\it impervious surfaces} and {\it roof} classes. 
The worst performance is for the {\it fence/hedge} and {\it powerline} classes, which according to Table \ref{tab:confusion} is due to the confusion between closely-related classes. 
For example, {\it powerline} is mainly confused with {\it roof}, and {\it fence/hedge} is confused with {\it shrub}, both of which have similar topological and spectral characteristics. 
Likewise, the accuracy of {\it low vegetation} is affected by the presence of {\it impervious surfaces} and {\it shrubs}. 
{\it Shrub} appears to be causing most of the confusion. 
This is likely due to the fact that the spectral information is similar among the vegetation classes {\it low vegetation}, {\it shrub}, and {\it trees}. 
While height information may improve the results, the presence of classes with similar heights such as {\it car} and {\it fence/hedge} make this differentiation challenging. 

To evaluate our performance against others, we compare our method to all submitted ISPRS contest results (both published and unpublished) at the time of paper submission. 
Since some submissions are unpublished we will review them briefly using the available information on the contest website\footnote{\url{https://goo.gl/6iTj6W}};  
We refer to the submitted methods according to names posted on the contest website. 
Interested readers are encouraged to review the website for further details. 

The {\bf IIS\_7}\footnote{\url{https://goo.gl/zepC6d}}  method used spectral and geometrical features, combined with a segmentation based on supervoxels and color-based region growing. 
In contrast to {\bf IIS\_7}, we don't handcraft geometrical features or utilize the spectral information to segment the point cloud into similar coherent regions before classification. 
The {\bf UM}\footnote{\url{https://goo.gl/uzZFcs}} method used a One-vs-One classifier based on handcrafted features, including point attributes such number of returns, textural properties, and geometrical attributes. 
The {\bf HM\_1}\footnote{\url{https://goo.gl/d7AmXY}} method utilized a CRF with RF classifier and contrast-sensitive Potts models, along with 2D-and 3D-geometrical features. 
The {\bf WhuY3}\footnote{\url{https://goo.gl/k7Qe1k}} deep learning method used a convolutional network operating on four features in a multi-scale fashion. 
The {\bf K\_LDA}~\citep{Blomley16} method used covariance features at multiple scales. 
Finally, the {\bf LUH}\footnote{\url{https://goo.gl/fGzrf3}} method used a two-layer hierarchical CRF that explicitly defines contextual relationships and utilizes voxel cloud connectivity segmentation, along with handcrafted features such as Fast Point Feature Histograms (FPFH).%{\mxy should we review the last 2??}

Per-class accuracy, and overall accuracy (OA) for each submission, including ours, are shown in Table \ref{tab:accuracy}. 
As seen in Table~\ref{tab:accuracy}, our method ranks second overall, sharing an overall accuracy of 81.6\% with {\bf LUH}.
The method with the highest overall accuracy is {\bf WhuY3}, achieving 82.3\%. 
However, {\bf WhuY3} achieved only one per-class highest accuracy score, as opposed to two for our method: the {\it car} and {\it shrub} classes.  
Likewise, the {\bf IIS\_7} method, which achieved the most (3) highest scores on a per-class basis, and use spectral data as well, only ranked eighth overall with an overall accuracy of 76.2\%. 
%was able to classify the {\it car} class better than other methods, while we share the highest accuracy with {\bf LUH} method for the {\it shrub} class. 
%While {\bf IIS\_7} had three high accuracy scores, {\bf UM} didn't lead on any of the classes. 

\begin{table*}[]
\centering
\caption{The per-class accuracy for each method and the corresponding Overall Accuracy (OA). Values in red correspond to the highest score, blue to second highest score, and green to third highest score.}
\label{tab:accuracy}
% \footnotesize
% \tabcolsep=0.11cm
\begin{tabular}{c|ccccccccc|l}
\textbf{Methods} & \textbf{power}  & \textbf{low\_veg}  & \textbf{imp\_surf} & \textbf{car} & \textbf{fence\_hedge} & \textbf{roof} & \textbf{fac} & \textbf{shrub} & \textbf{tree} & \textbf{OA}   \\ \hline
\textbf{Ours}   & 29.8   & 69.8    & 93.6   & {\color[HTML]{FE0000} \textbf{77.0}} & 10.4  & 92.9   & 47.4   & {\color[HTML]{FE0000} \textbf{73.4}} & 79.3   & {\color[HTML]{3531FF} \textbf{81.6}} \\
\textbf{IIS\_7} & 40.8  & 49.9 & {\color[HTML]{FE0000} \textbf{96.5}} & 46.7 & {\color[HTML]{FE0000} \textbf{39.5}} & {\color[HTML]{FE0000} \textbf{96.2}} & --- & 52.0 & 68.8 & 76.2                \\
\textbf{UM}     & 33.3  & {\color[HTML]{3531FF} \textbf{79.5}}  & 90.3 & 32.5 & 02.9 & 90.5 & 43.7  & 43.3    & {\color[HTML]{FE0000} \textbf{85.2}} & {\color[HTML]{009901}\textbf{80.8}}          \\
\textbf{HM\_1}  & {\color[HTML]{3531FF} \textbf{82.8}} & 65.9 & {\color[HTML]{3531FF} \textbf{94.2}} & 67.1 & 25.2 & 91.5 & 49.0 & {\color[HTML]{3531FF} \textbf{62.7}} & {\color[HTML]{3531FF} \textbf{82.6}} & 80.5 \\
\textbf{WhuY3}  & 0.247 & {\color[HTML]{FE0000} \textbf{81.8}} & 91.9 & {\color[HTML]{3531FF} \textbf{69.3}} & 14.7 & {\color[HTML]{3531FF} \textbf{95.4}} & 40.9 & 38.2 & 78.5 & {\color[HTML]{FE0000} \textbf{82.3}} \\
\textbf{K\_LDA} & {\color[HTML]{FE0000} \textbf{89.3}} & 12.4 & 47.6 & 28.9 & 20.4 & 80.7 & {\color[HTML]{3531FF} \textbf{51.3}} & 38.4 & 72.8 & 50.2                          \\
\textbf{LUH} & 53.2 & 72.7 & 90.4 & 63.3 & {\color[HTML]{3531FF} \textbf{25.9}} & 91.3 & {\color[HTML]{FE0000} \textbf{60.9}} & {\color[HTML]{FE0000} \textbf{73.4}} & 79.1 & {\color[HTML]{3531FF} \textbf{81.6}} \\
%\textbf{BIJ\_W}          & 0.077                                 & 0.800                                 & 0.925                                 & 0.521                                 & {\color[HTML]{3531FF} \textbf{0.269}} & 0.918                                 & 0.386                                 & 0.484                                 & 0.769                                 & 0.815                                 \\
%\textbf{NANJ}            & 0.683                                 & {\color[HTML]{3531FF} \textbf{0.804}} & 0.920                                 & {\color[HTML]{3531FF} \textbf{0.726}} & ---                                   & {\color[HTML]{3531FF} \textbf{0.955}} & 0.306                                 & ---                                   & {\color[HTML]{FE0000} \textbf{0.894}} & 0.807                                 \\ 
\cline{1-11}
\end{tabular}
\end{table*}

\begin{table*}[]
\centering
\caption{The F1-scores per-class for each method and the average value. Values in red correspond to the highest score, and blue to second highest score. Our score is marked with green color.}
\label{tab:F1}
% \footnotesize
% \tabcolsep=0.11cm
\begin{tabular}{c|ccccccccc|l}
                        \textbf{Methods} & \textbf{power}                        & \textbf{low\_veg}                     & \textbf{imp\_surf}                    & \textbf{car}                          & \textbf{fence\_hedge}                 & \textbf{roof}                         & \textbf{fac}                          & \textbf{shrub}                        & \textbf{tree}                         & \textbf{Avg. F1}                       \\ \cline{1-11} 
\multicolumn{1}{c|}{\textbf{Ours}}   & 37.5                                 & 77.9                                 & {\color[HTML]{FE0000} \textbf{91.5}} & {\color[HTML]{FE0000} \textbf{73.4}} & 18.0                                 & {\color[HTML]{3531FF} \textbf{94.0}} & 49.3                                 & 45.9                                 & {\color[HTML]{3531FF} \textbf{82.5}} & {\color[HTML]{009901} \textbf{63.33}} \\
\multicolumn{1}{c|}{\textbf{IIS\_7}} & 54.4                                 & 65.2                                 & 85.0                                 & 57.9                                 & 28.9                                 & 90.9                                 & ---                                   & 39.5                                 & 75.6                                 & 55.27                                 \\
\multicolumn{1}{c|}{\textbf{UM}}     & 46.1                                 & {\color[HTML]{3531FF} \textbf{79.0}} & 89.1                                 & 47.7                                 & 05.2                                 & 92.0                                 & 52.7                                 & 40.9                                 & 77.9                                 & 58.96                                 \\
\multicolumn{1}{c|}{\textbf{HM\_1}}  & {\color[HTML]{FE0000} \textbf{69.8}} & 73.8                                 & {\color[HTML]{FE0000} \textbf{91.5}} & 58.2                                 & {\color[HTML]{3531FF} \textbf{29.9}}                                 & 91.6                                 & {\color[HTML]{3531FF} \textbf{54.7}} & {\color[HTML]{FE0000} \textbf{47.8}} & 80.2                                 & {\color[HTML]{3531FF} \textbf{66.39}} \\
\multicolumn{1}{c|}{\textbf{WhuY3}}  & 37.1                                 & {\color[HTML]{FE0000} \textbf{81.4}} & 90.1                                 & 63.4                                 & 23.9                                 & 93.4                                 & 47.5                                 & 39.9                                 & 78.0                                 & 61.63                                 \\
\multicolumn{1}{c|}{\textbf{K\_LDA}} & 05.9                                 & 20.1                                 & 61.0                                 & 30.1                                 & 16.0                                 & 60.7                                 & 42.8                                 & 32.5                                 & 64.2                                 & 37.03                                 \\
\multicolumn{1}{c|}{\textbf{LUH}}    & {\color[HTML]{3531FF} \textbf{59.6}}                                 & 77.5                                 & {\color[HTML]{3531FF} \textbf{91.1}} & {\color[HTML]{3531FF} \textbf{73.1}} & {\color[HTML]{FE0000} \textbf{34.0}} & {\color[HTML]{FE0000} \textbf{94.2}} & {\color[HTML]{FE0000} \textbf{56.3}} & {\color[HTML]{3531FF} \textbf{46.6}} & {\color[HTML]{FE0000} \textbf{83.1}} & {\color[HTML]{FE0000} \textbf{68.39}} \\
%\multicolumn{1}{c|}{\textbf{BIJ\_W}} & 0.138                                 & 0.785                                 & 0.905                                 & 0.564                                 & {\color[HTML]{FE0000} \textbf{0.363}} & 0.922                                 & 0.532                                 & 0.433                                 & 0.784                                 & 0.6029                                 \\
%\multicolumn{1}{c|}{\textbf{NANJ}}   & {\color[HTML]{3531FF} \textbf{0.601}} & 0.777                                 & 0.909                                 & 0.517                                 & ---                                   & 0.936                                 & 0.338                                 & ---                                   & 0.771                                 & 0.5388                                 \\ 
\cline{1-11}
\end{tabular}
\end{table*} 

We also evaluated our results using the F1-score, which is generally considered to be a better comparative metric when an uneven class distribution exists and/or when the costs of false positives and false negatives are very different.
%In this application, the former condition is true while the latter condition is not necessarily the case if the separation of all classes is considered the task at hand.
Table~\ref{tab:F1} compares our method to others using the F1-score. Our method performed well across all classes except for the {\it fence/hedge} class. Other methods demonstrated similarly poor results on this same class. While {\bf LUH} did score the highest on the {\it roof} and {\it tree} classes, their scores were only marginally better than ours, 0.2\% and 0.6\%, respectively. Their higher performance for the {\it fence/hedge} and {\it powerline} classes did, however, allow them to achieve the highest F1-score of 68.39\%, with {\bf HM\_1} ranking second with a score of 66.39\%. Our presented technique did ranked third with a score of 63.33\%, surpassing {\bf WhuY3} which scored the highest on overall accuracy (Table~\ref{tab:accuracy}).

% Moved to previous paragraph
%The overall accuracy for {\bf IIS\_7} is 76.2\% compared to ours \%81.6. 
%We share the same overall accuracy and the second rank with {\bf LUH}. 
The scores in both Tables should be thought of in context of the algorithm complexity. 
In other words, minimal accuracy gains may not be worth the added computational overhead. 
The {\bf LUH} method, for example, fares well in both overall, and per-class, accuracy and F1-scores. 
%we learn the point-wise features and the per-block contextual descriptors using 1D-convolutions and a simple pooling layer without operating on structured or segmented regions. 
%The method {\bf WhuY3} scored the highest overall accuracy of \%82.3. 
However, this method uses two independent CRFs with handcrafted features, and segmentation methods that force points within a segment to share the same label. Likewise, {\bf HM\_1} uses a variety of contrast-sensitive Potts models, which may help preserve edges during segmentation, but adds NP-hard components to the problem~\citep{potts_model_2001}. As a result, while smoother results may be achieved for small data sets, this comes at the cost of slow run-time performance (see Section\ref {sec:direct_methods}) and scalability limitations for massive data sets. 
%Since the authors only describe four handcrafted features, we think they transformed the features into image-like maps and the used CNN is a 2D-network compared to ours which 
In contrast, we utilize a series of simple 1D-convolutions that operate directly on the point cloud, without engineering additional features, or requiring structured representations such as segments. 
Instead point-wise features (\eg~ spatial coordinates and/or spectral information), and per-block contextual descriptors are learned in a straightforward, end-to-end fashion. Our average inference time is 3.7s for a point cloud with $N$ approximately equal to 412k. %using 1D-convolutions and a simple pooling layer without operating on structured or segmented regions. 
%Looking at the results from a different prospective, Table~\ref{tab:F1} compare our method to others using the F1-score. While  

\subsection{Effect of Individual Features}
\label{sec:effect_of_individual_features}
Our 1D convolutional network has flexible input feature requirements, and can consume directly (i) spatial coordinates only, \ie~ a 3D point cloud, (ii) spatial coordinates and spectral information, or (iii) spectral information only. %, as will be shown in Section \ref{sec:extension_to_2D} for 2D semantic labeling. 
Moreover, 3D spatial coordinates (x,y,z) may optionally be normalized to remove the effects of terrain, providing (x,y,height-above-ground). 
Finally, models can be trained at different scales by adjusting the block size. 
In this section we provide two experiments to analyze the impact of feature selection and digital terrain model on model performance for various use-cases. 

\begin{figure*}[th]
\begin{center}
\includegraphics[width=15cm]{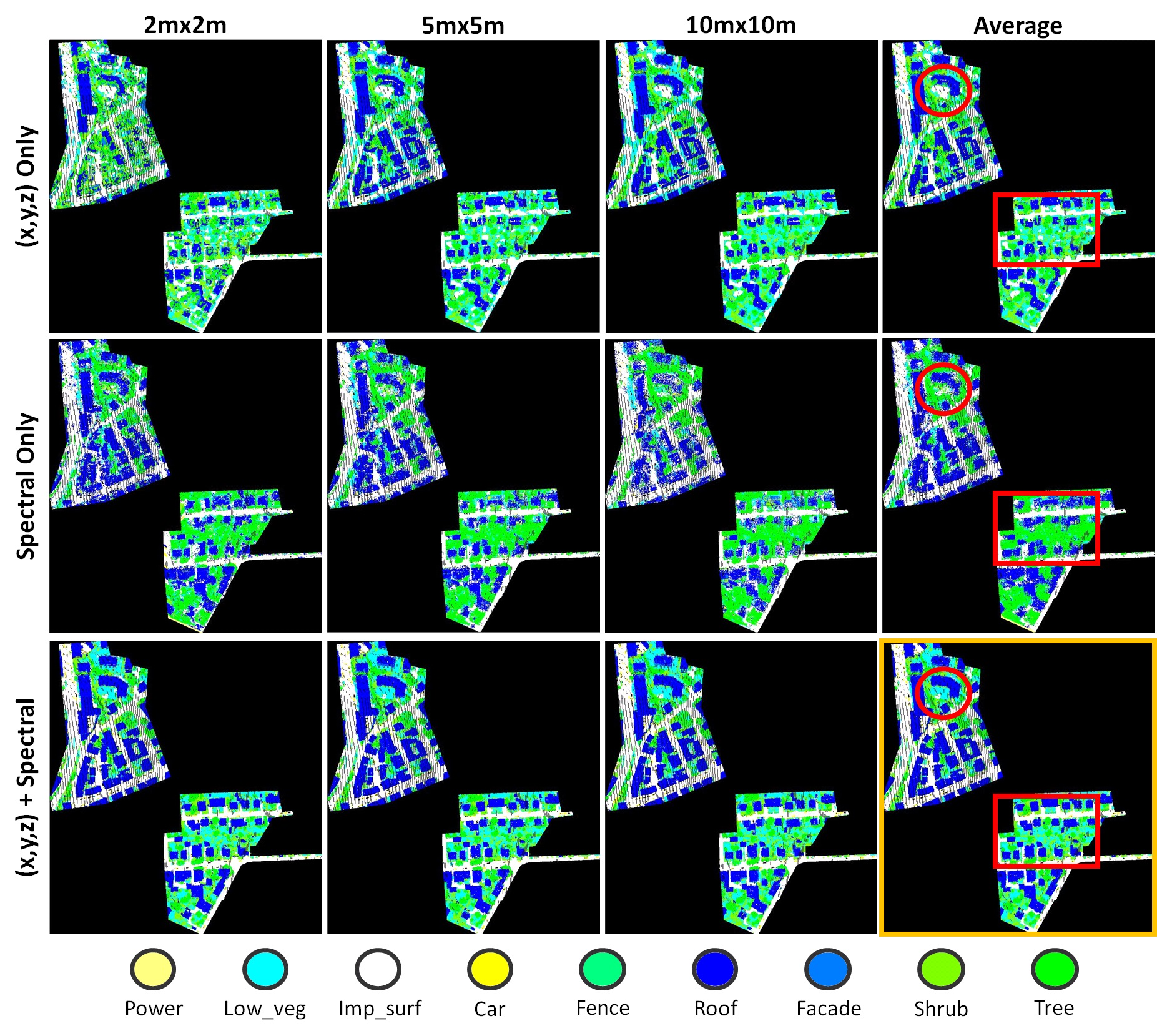}
\end{center}
\caption{Results matrix showing the effect of training the network using different input features (rows), and at different block sizes (columns). The submitted result to the ISPRS 3D Semantic Labeling Contest is shown in the bottom right. Red markers indicate regions of comparison. Class color keys are shown at the bottom.}
\label{fig:exp1}
\end{figure*}

In the first experiment, we investigate the effect of the input feature selection, \ie~ spatial and/or spectral information, by training our model based on three sets of input data.  
The first model is trained using 3D coordinates only. 
The second model is trained using spectral information only.
The third model is trained using both 3D-coordinates and spectral information (IR-R-G) for each point; this is the best-performing network, which was evaluated in more detail in Section \ref{sec:classification_results}.
Results are shown in Figure \ref{fig:exp1} for multiple scales, \ie~  block sizes of 2m$\times$2m, 5m$\times$5m, and 10m$\times$10m, and,  in column 4, the average result across all scales.
%only, while the last model is our main network that uses all features. 
%The corresponding multi-scale results are shown in Figure \ref{fig:exp1}. 
%The rows correspond to the features used during training, while the columns indicate the used scales. 
%The last column shows the average result across all scales. 

When using the 3D-coordinates only (first row), the larger scale performs better than the smaller scale.
On the other hand, when using the spectral information only (second row), the smaller scale performs better than the larger scale. 
This result is interesting since it shows the effect of global features at multiple scales. 
For example, when using the spectral data, smaller scales will generally include similar points as opposed to larger scales. 
Therefore, the global features will accurately describe the group (block) of points at a smaller scale. 
In contrast, the global features will not sufficiently describe structures when using only the 3D-coordinates at a smaller scale. 
In this case, a larger scale is needed to capture structural information that can help distinguish between 3D objects.

This suggests that combining both features could improve the results. 
%Therefore, the last row shows our results after averaging multiple scales.
The result submitted to the ISPRS 3D Semantic Labeling Contest (bottom right corner) used the average of 3 block sizes, trained on both (x,y,z) point coordinates and spectral information. 
%Our results at each scale is smoother and better compared to the first and the second rows. 
The red circle highlights how using all features helped to correctly classify a troublesome {\it low vegetation} region. 
Likewise, the red box highlights how the spectral data, in the absence of 3D coordinates, tends to classify all vegetation as {\it tree}, while the 3D-coordinates, in the absence of spectral data tends to confuse {\it impervious surfaces} with {\it low vegetation}.
See Figure \ref{fig:results&rg} for reference regrading correctly classified regions in our submission.

\begin{figure*}[th]
\begin{center}
\includegraphics[width=16cm]{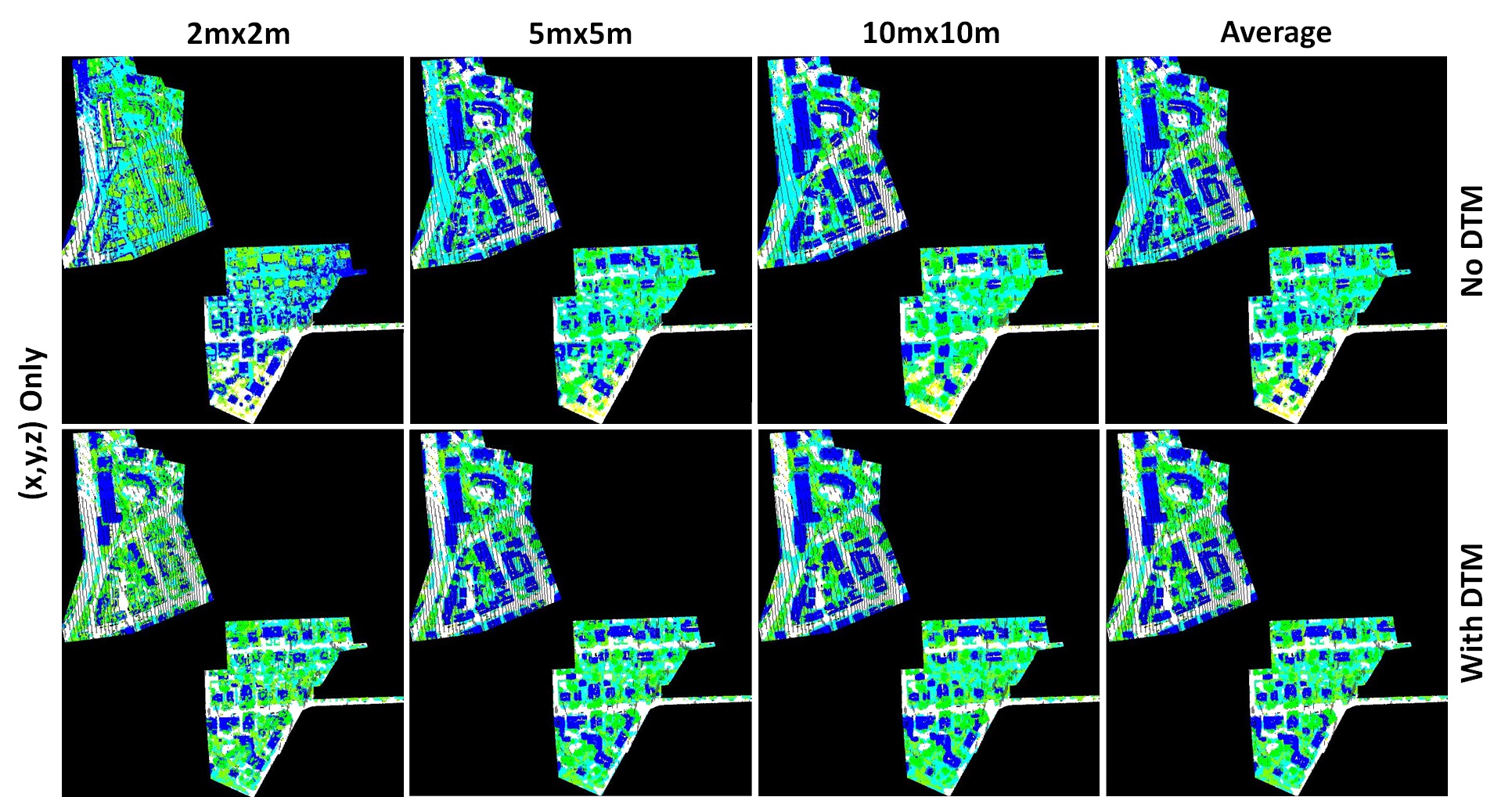}
\end{center}
\caption{A comparison between training with (bottom) and without (top) a digital terrain model using only 3D-coordinates for block sizes. }
\label{fig:exp_xyz}
\end{figure*}

\begin{figure*}[th]
\begin{center}
\includegraphics[width=16cm]{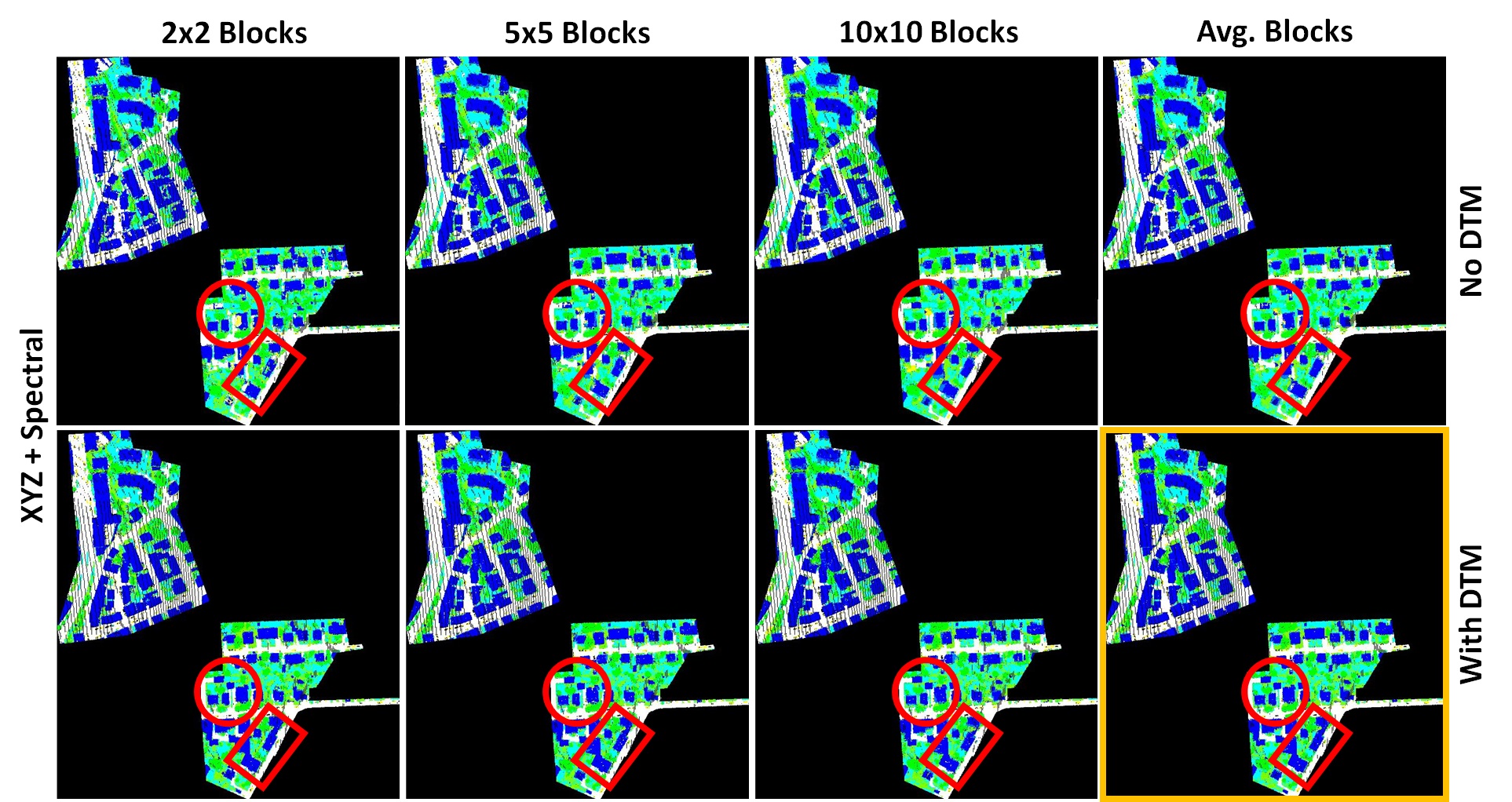}
\end{center}
\caption{A comparison between training with and without a digital terrain model using 3D-coordinates and spectral data. Regions marked with red highlight differences.}
\label{fig:exp_xyzrgb}
\end{figure*}

In the second experiment, we investigated the effect of normalizing the z-coordinates to height-above-ground, based on the DTM model. 
This was done both in the absence (Figure \ref{fig:exp_xyz}) and presence (Figure \ref{fig:exp_xyzrgb}) of spectral information, which may not always be available.  
%Results of training our network using 3D-coordinates only are shown in Figure \ref{fig:exp_xyz}, while Figure \ref{fig:exp_xyzrgb} shows the results when training using all features. 
Figure \ref{fig:exp_xyz} shows that in the absence of spectral information, normalizing the 3D coordinates to height-above-ground provides a cleaner and less-fragmented classification at all scales, including the average.
%See error map in Figure \ref{fig:results&rg} for reference.  

%As shown in Figure 7, the best classification results are achieved when spectral information is available, and after normalizing to adjust for the DTM. 
As shown in Figure \ref{fig:exp_xyzrgb}, the best classification results are achieved when spectral information is available, and after obtaining height-above-ground using a DTM.
Close scrutiny of the regions marked in red shows that terrain-normalized input points improves the classification, especially for parts of roofs. 
However, in general, when spectral information is available, the results without using a DTM are still reasonable.
This is exciting, as it opens the door to more streamlined point cloud exploitation workflows that can directly ingest 3D data in its original form.
Furthermore, this may enable new techniques for generating, at scale, more precise DTMs based on the spectral content inherent in stereo-derived point clouds \citep{tapper2016extraction}.

\subsection{Extension to 2D Semantic Labeling}
\label{sec:extension_to_2D}
A primary contributions of our method is the compact, 1D CNN architecture.
This complements the input data characteristics of point clouds in ways that traditional 2D CNNs do not. 
In other words, deep learning methods that rely on mapping features to 2D image-like maps, \ie~ DSM or density maps~\citep{MVCNN,Luca,Yansong} don't take into account the increased complexity when adding new features. 
In such a case, adding a new feature involves concatenating the data with a new (full) 2D-channel, subsequently requiring an additional set of 2D-filters and greatly increasing the amount memory required. 
On the other hand, our method operates directly on the point cloud, representing each point with a 1D-vector. 
Adding a new feature only requires appending each per-point vector with a single value; this increases the width of the 1D-convolutions by only one element in the first layer and does not modify the rest of the network. 
This advantage provides an elegant solution to the 3D multi-sensor fusion problem, allowing us to readily combine complementary information about the scene from different modalities for semantic segmentation. 

Additionally, we show that this 1D architecture can be easily extended to the traditional task of 2D-image labeling. 
%Here we show how our method can be extended handle different modalities utilizing the Potsdam 2D-Semantic Labeling contest\footnote{\url{https://goo.gl/rN3Cge}} as a case study. 
To handle 2D images, a preprocessing step simply restructures the data from a raster representation to a 2D-array where each row represents a 3D-point vector.% as shown in Figure \ref{fig:2D_seg_upper}. 
The spatial position of each point corresponds to the 2D-pixel coordinates, while the height value corresponds to the digital counts in the DSM image.
The spectral information of each point is derived from the corresponding image's digital numbers. . 
%The upper part of Figure \ref{fig:2D_seg} shows an example of transforming a spectral image and a DSM channel to an array where each row represent a single point with the corresponding feature vector. 
We then train our model as described previously, excluding the data augmentation or multi-scale stages, due to the high resolution nature of the images which provided sufficient training samples. %which are unnecessary for the 2D-semantic classification task.
\begin{figure*}[th]
\begin{center}
\includegraphics[width=16cm]{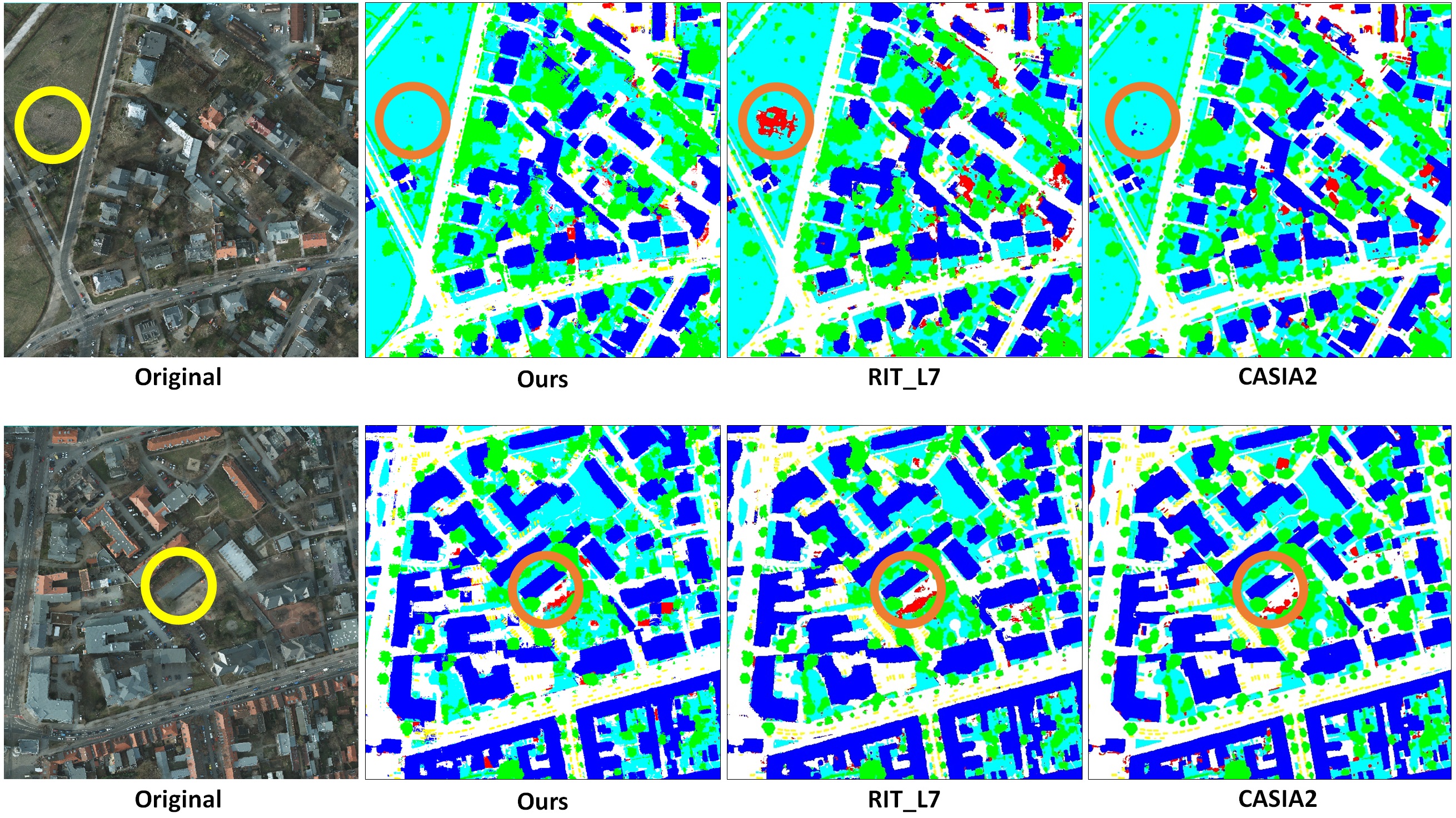}
\end{center}
\caption{A qualitative comparison between our method to two submissions that uses 2D-deep networks. The yellow circles shows the original regions of interest, while the brown circles mark the corresponding regions in the classification maps.}
\label{fig:2D_seg_lower}
\end{figure*}

We qualitatively evaluated our method against two relevant submissions from the Potsdam 2D-Semantic Labeling Contest\footnote{\url{https://goo.gl/rN3Cge}}. 
%{\mxy Quantitative results are unavailable at the moment since we didn't participate in the challenge.}
The first method, {\bf RIT\_L7}~\citep{Yansong} used a CRF model with an FCN and a logistic regression classifier to fuse the normalized DSM and the spectral information, scoring an overall accuracy of 88.4\%.
The {\bf RIT\_L7} method was chosen for comparison since it uses CRF to explicitly utilize contextual information.
%It should be noted that the original LiDAR point cloud was available to the participants. 
The second method (unpublished), {\bf CASIA2}, fine-tuned Resnet101~\citep{resnet} and used only the spectral data, scoring an overall accuracy of 91.1\%. 
 The {\bf CASIA2} method was chosen for comparison since it relies on a very large and computationally intensive Network, in contrast to our very compact network. 
For example, our network has about 1.9M parameters, while Resnet101 has about 44.5M parameters. 

Figure \ref{fig:2D_seg_lower} compares our output semantic labeling results against the {\bf RIT\_L7} and {\bf CASIA2} methods, for two example 2D images. 
%The first example shows an area dominated  low vegetation regions. 
% \begin{figure}[t]
% \begin{center}
% \includegraphics[width=7.7cm]{images/2D_convert_upper_v2.png}
% \end{center}
% \caption{The upper part shows how our method can easily be extended to handle 2D-semantic segmentation by simple modify the data representation from image-like to an array of point clouds. The lower part qualitatively compares our method to two submissions that uses 2D-deep networks.}
% \label{fig:2D_seg_upper}
% \end{figure}
The circled areas in the upper image show that our method was able to correctly (see the corresponding spectral image) classify challenging low-vegetation regions, which were incorrectly classified as clutter by {\bf RIT\_L7}, and had residuals of the building class in {\bf CASIA2}. 
%The second example shows a mostly residential area with different building shapes and types. 
Likewise, the circled regions in the lower image shows that both our method and the {\bf RIT\_L7} method produced good results when classifying the center building. 
The {\bf CASIA2} method missed the top right corner of the building; this is likely due to a lack of consideration for height information. 
Including height information as another channel in a very large model such as Resnet101 is not a trivial task due to three-channel design.  
Also, even if height information were included as a fourth channel, finetuning would not be possible and training would be infeasible given the limited number of images provided in the contest. 
This highlights the advantages of our compact model: adding another feature is as simple as adding a single value per-point. 
And, given the relatively few number of parameters, a simple data augmentation is sufficient to train our network.
%\begin{figure*}[th]
%\begin{center}
%\includegraphics[width=17cm]{images/2D_convert.png}
%\end{center}
%\caption{The upper part shows how our method can easily be extended to handle 2D-semantic segmentation by simple modify the data representation from image-like to an array of point clouds. The lower part qualitatively compares our method to two submissions that uses 2D-deep networks.}
%\label{fig:2D_seg}
%\end{figure*}
% \begin{figure*}[th]
% \begin{center}
% \includegraphics[width=10cm]{images/No_DEM_Spectral_and_XYZ_Sectral&XYZ.png}
% \end{center}
% \caption{The basic semantic labeling network takes an Nx3 set of point cloud, and pass it through a series of convolutional layers. The upper part shows the point level and the global feature learning stage. The global features are found through a symmetric max pooling layer applied across the points. The lower part shows a the classification stage where concatenated features are passed though a $(1\times1)$ convolutional layers then to a softmax classifier. }
% \label{fig:test1}
% \end{figure*}

\section{Conclusions}
\label{sec:conclusions}
In this paper we present a deep learning framework to semantically label 3D point clouds with spectral information. 
Our compact, 1D fully convolutional architecture directly consumes unstructured and unordered point clouds without relying on costly engineered features or 2D image transformations. 
We achieved near state-of-the-art results using only the 3D-coordinates and their corresponding spectral information. 
By design, our network can consume regions with varying densities, and is able to learn local and global features in an end-to-end fashion. 
Furthermore, our model is flexible, and can be readily extended to 2D semantic segmentation. 
Also, our experiments showed promising results when classifying unnormalized points. 
Given the compact, end-to-end framework, and fast testing time, our model has the potential to scale to much larger datasets, including those derived from optical satellite imagery.
%think that the semantic understanding of 3D point cloud can be easily incorporated in larger frameworks when minimal modifications are applied. 
Future work will extend the model to operate on optically derived point clouds, improve the performance with respect to  unnormalized points, and investigate more fine-grained classes. 
%Also, we are investigating the addition of region-wise layers where multiple regional features could be learned within a single block instead of a global block-wise feature.

\section*{Acknowledgments}
This material is based upon work supported by the U.S. Department of Energy, Office of Science. 

\section*{Copyright}
This manuscript has been authored by one or more employees of UT-Battelle, LLC under Contract No. DE-AC05-00OR22725 with the U.S. Department of Energy. The United States Government retains and the publisher, by accepting the article for publication, acknowledges that the United States Government retains a non-exclusive, paid-up, irrevocable, world-wide license to publish or reproduce the published form of this manuscript, or allow others to do so, for United States Government purposes. The Department of Energy will provide public access to these results of federally sponsored research in accordance with the DOE Public Access Plan (http://energy.gov/downloads/doe-public-access-plan).
\section*{References}

%% The Appendices part is started with the command \appendix;
%% appendix sections are then done as normal sections
%% \appendix

%% \section{}
%% \label{}

%% If you have bibdatabase file and want bibtex to generate the
%% bibitems, please use
%%
\bibliographystyle{elsarticle-harv} 
\bibliography{library}

%% else use the following coding to input the bibitems directly in the
%% TeX file.

\end{document}